\documentclass{article}
\pdfoutput=1

\usepackage[final, nonatbib]{neurips_2022}

\usepackage{microtype}
\usepackage{graphicx}
\usepackage{subcaption}
\usepackage{booktabs} 

\usepackage{float} 

\newcommand{\zzhao}[1]{{\color{black} #1}}
\newcommand{\change}[1]{{\color{black} #1}}
\newcommand{\nipsedit}[1]{{\color{black} #1}}
\newcommand{\zzhaocamera}[1]{{\leavevmode\color{black} #1}}
\newcommand{\cliucamera}[1]{{\leavevmode\color{black} #1}}

\newcommand{\splitcell}[1]{\begin{tabular}{@{}c@{}}#1\end{tabular}}
\newcommand{\bsplitcell}[1]{$\left[ \splitcell{#1} \right]$}

\usepackage[utf8]{inputenc} 
\usepackage[T1]{fontenc}    
\usepackage{url}            
\usepackage{hyperref}
\usepackage{booktabs}       
\usepackage{amsfonts}       
\usepackage{nicefrac}       
\usepackage{microtype}      
\usepackage{xcolor}         
\usepackage{wrapfig}

\usepackage{amsmath}
\usepackage{amssymb}
\usepackage{mathtools}
\usepackage{amsthm}
\usepackage{multirow}
\usepackage{multicol}
\usepackage{makecell}
\usepackage{xcolor}
\usepackage{arydshln}
\usepackage[utf8]{inputenc}
\usepackage{algorithm}
\usepackage{algorithmic}

\usepackage[capitalize,noabbrev]{cleveref}

\theoremstyle{plain}
\newtheorem{theorem}{Theorem}[section]

\theoremstyle{definition}

\theoremstyle{remark}

\usepackage[textsize=tiny]{todonotes}

\title{Robust Binary Models by Pruning Randomly-initialized Networks}

\author{%
    Chen Liu$^*$~~~~Ziqi Zhao\thanks{indicates equal contributions}~~~~Sabine S\"usstrunk~~~~Mathieu Salzmann\\
    École Polytechnique Fédérale de Lausanne, Lausanne, Switzerland, \\
    \texttt{\{chen.liu, ziqi.zhao, sabine.susstrunk, mathieu.salzmann\}@epfl.ch}
}

\newcommand{\defeq}{\mathrel{\mathop:}=}

\def\eqref#1{equation~\ref{#1}}

\def\1{\bm{1}}

\def\rvb{{\mathbf{b}}}

\def\rvg{{\mathbf{g}}}

\def\rvm{{\mathbf{m}}}

\def\rvo{{\mathbf{o}}}

\def\rvs{{\mathbf{s}}}

\def\rvu{{\mathbf{u}}}
\def\rvv{{\mathbf{v}}}
\def\rvw{{\mathbf{w}}}
\def\rvx{{\mathbf{x}}}

\def\rmW{{\mathbf{W}}}

\def\gD{{\mathcal{D}}}

\def\gL{{\mathcal{L}}}
\def\gM{{\mathcal{M}}}

\def\gS{{\mathcal{S}}}

\def\sR{{\mathbb{R}}}

\begin{document}

\maketitle

\begin{abstract}
Robustness to adversarial attacks was shown to require a larger model capacity, and thus a larger memory footprint. In this paper, we introduce an approach to obtain robust yet compact models by pruning randomly-initialized binary networks.
Unlike adversarial training, which learns the model parameters, we initialize the model parameters as either $+1$ or $-1$, keep them fixed, and find a subnetwork structure that is robust to attacks.
Our method confirms the \textit{Strong Lottery Ticket Hypothesis} in the presence of adversarial attacks, and extends this to binary networks.
Furthermore, it yields more compact networks with competitive performance than existing works by 1) adaptively pruning different network layers; 2) exploiting an effective binary initialization scheme; 3) incorporating a last batch normalization layer to improve training stability.
Our experiments demonstrate that our approach not only always outperforms the state-of-the-art robust binary networks, but also can achieve accuracy better than full-precision ones on some datasets.
Finally, we show the structured patterns of our pruned binary networks.
\end{abstract}

\section{Introduction}

Deep neural networks have achieved unprecedented success in machine learning~\cite{dosovitskiy2021an, he2016deep, vaswani2017attention}.
However, their state-of-the-art performance comes with costs.
First, modern deep neural networks usually have millions of parameters, making them difficult to deploy on devices with limited memory or computational power.
Second, these models are vulnerable to adversarial attacks: imperceptible perturbations of the input can dramatically change their output and lead to incorrect predictions~\cite{szegedy2013intriguing}.
Furthermore, jointly addressing both issues is complicated by the fact that, as shown in~\cite{madry2017towards, xie2020intriguing}, achieving robustness against adversarial attacks typically requires higher network capacity.

In this paper, we introduce an approach to obtaining compact and robust binary neural networks.
Our method follows a fundamentally different philosophy from typical adversarial training~\cite{madry2017towards}: instead of using adversarial examples to train the model parameters, we fix the model parameters and search for a robust network structure.
To this end, and to simultaneously achieve compactness, we prune randomly-initialized binary networks.
The resulting sparse and binary networks have much smaller memory footprint than the dense or full-precision ones.
They are inherently lightweight and robust.


Our work is motivated by the \textit{Strong Lottery Ticket Hypothesis}~\cite{ramanujan2020s}, which observed that within a random overparameterized network, there exists a subnetwork achieving performance similar to that of trained networks with the same number of parameters.
\change{The \textit{Robust Scratch Ticket}~\cite{fu2021drawing} extends this hypothesis to the context of adversarial robustness. Here, we introduce a novel pruning strategy that yields higher compression rates, and investigate the case of binary model parameters.
Specifically, we develop an \textit{adaptive pruning strategy} to adaptively use different pruning rates for different layers.
Furthermore, we introduce a normalization-based technique to increase the algorithm's stability for binary networks.
The subnetworks obtained by our method are consequently more compact than a full-precision one, while achieving a similar robustness to attacks.
}

We conduct extensive experiments on standard benchmarks to confirm the effectiveness of our method.
We obtain both better performance and a more stable training behavior than existing works~\cite{fu2021drawing}. Furthermore, our approach outperforms the state-of-the-art robust binary networks~\cite{gui2019model}, achieving performance on par with or even better than the state-of-the-art robust full-precision ones\zzhao{~\cite{chen2022sparsity, ozdenizci2021training, wang2020hydra}} while producing much more compact networks.

Finally, we conduct preliminary investigations on the structure of the robust subnetworks obtained by our algorithm.
\change{We find our methods prefer to prune the whole channel or the kernel in the convolutional layers.
In addition, for two consecutive convolutional layers, the kernels pruned in the first layer are well aligned with the channels pruned in the second layer.
Altogether, our work sheds some light on understanding the structure of robust networks of high parameter sparsity, it also indicates the potential of regular pruning.}

Our code is available at \href{https://github.com/IVRL/RobustBinarySubNet}{https://github.com/IVRL/RobustBinarySubNet}.

\textbf{Notation.} We use light letters, lowercase bold letters, and uppercase bold letters to represent scalars, vectors, and higher dimensional tensors, respectively.
$\odot$ is the elementwise multiplication operation.
We use the term \textit{adversarial budget} to represent the range of allowable perturbations.
Specifically, the adversarial budget $\gS_\epsilon$ is based on the $l_\infty$ norm and defined as $\{\Delta | \|\Delta\|_\infty \leq \epsilon\}$, with $\epsilon$ the strength of the adversarial budget.
We refer to the proportion of pruned parameters over the total number of parameters in a layer or a model as the \textit{pruning rate} $r$.

\section{Related Work}

\textbf{Adversarial Robustness.} Deep neural networks have been shown to be vulnerable to adversarial attacks~\cite{szegedy2013intriguing, moosavi2017universal}.
To generate adversarial examples, the \textit{Fast Gradient Sign Method} (FGSM)~\cite{goodfellow2014explaining} perturbs the input in the direction of the input gradient.
The \textit{Iterative Fast Gradient Sign Method} (IFGSM)~\cite{kurakin2016adversarial} improves FGSM by running it iteratively.
\textit{Projected Gradient Descent} (PGD)~\cite{madry2017towards} uses random initialization and multiple restarts on top of IFGSM to further strengthen the attack.
Recently, AutoAttack (AA)~\cite{croce2020reliable} has led to  state-of-the-art attacks by ensembling different types of attacks; it is used to reliably benchmark the robustness of models~\cite{croce2020robustbench} and we thus use it in our experiments.

Many works have proposed defense mechanisms against these adversarial attacks. Early ones~\cite{buckman2018thermometer, pang2020rethinking, xiao2020enhancing} used obfuscated gradients~\cite{athalye2018obfuscated, croce2020reliable} and thus were ineffective against adaptive attacks.
As a consequence, adversarial training~\cite{madry2017towards} and its variants~\cite{alayrac2019labels, carmon2019unlabeled, hendrycks2019using, rebuffi2021data, sinha2019harnessing, wu2020adversarial, zhang2019you, zhang2019theoretically} have in practice become the mainstream approach to obtain robust models.
Specifically, given a dataset $\{(\rvx_i, y_i)\}_{i = 1}^N$, a model $f$ parameterized by $\rvw$ and a loss function $\gL$, adversarial training solves the min-max optimization problem: 
\begin{equation}
\begin{aligned}
\min_{\rvw} \frac{1}{N} \sum_{i = 1}^N \max_{\Delta_i \in \gS_\epsilon} \gL(f(\rvw, \rvx_i + \Delta_i), y_i)\;.
\end{aligned} \label{eq:minmax}
\end{equation}
In practice, this is achieved by first generating adversarial examples $\rvx_i + \Delta_i$, usually by PGD, and then using these examples to train the model parameters.

While effective, adversarial training was shown to require a larger model capacity~\cite{madry2017towards, xie2020intriguing}.
Specifically, as the model capacity decreases, adversarial training first fails to converge while the training on clean inputs still yields non-trivial performance.
Conversely, as the model capacity increases, the performance of training on clean inputs saturates before that of adversarial training.
This highlights the challenge of finding robust yet compact models.
Here, we introduce a solution to this problem.

\textbf{Model Compression.} There are many ways to compress deep neural networks to achieve lower memory consumption and faster inference, including pruning, quantization, and parameter encoding.
The pioneering works used information-theoretic methods~\cite{lecun1990optimal} or second-order derivatives~\cite{hassibi1993optimal} to compress models by removing unimportant weights.
The seminal work~\cite{han2015learning} proposed to prune the parameters with the smallest absolute values for deep networks.
This motivated many follow-up works, performing either irregular pruning~\cite{guo2016dynamic, yang2017designing, zhang2018systematic}, which removes individual parameters, or regular pruning~\cite{he2017channel, liu2017learning}, which aims to discard entire convolutional kernels.
In contrast to pruning, quantization~\cite{zhou2016incremental, polino2018model, jin2020adabits} seeks to reduce the memory consumption and inference time by using low-precision parameters.
An extreme case of quantization is binarization, which can take the form of only binarizing the parameters~\cite{courbariaux2015binaryconnect, courbariaux2016binarized} or binarizing both parameters and intermediate activations~\cite{hubara2016binarized}.
The models can be further compressed by combining pruning with quantization and Huffman coding~\cite{han2015deep}.

Recently, some efforts have been made to incorporate adversarial training into model compression.
\cite{galloway2018attacking, lin2018defensive, rakin2018defend} suggest quantization as a defense against adversarial attacks.
\change{\cite{ozdenizci2021training} uses Bayesian connectivity sampling to prune the network while preserving its robustness.
\cite{chen2022sparsity} dynamically generates a robust subnetwork during adversarial training.}
\cite{ye2019adversarial} uses the alternating direction method of multipliers (ADMM) to alternatively conduct adversarial training and network pruning.
\cite{gui2019model} extends this framework 
to include other model compression techniques, such as quantization.
Furthermore,~\cite{wang2020hydra} introduces the HYDRA framework, which improves the performance of compressed robust models by a three-phase method: Pretraining, score-based pruning, and fine-tuning.
Here, we follow a different strategy: Instead of performing adversarial training, we search for a robust binary subnetwork in a randomly-initialized one.
We show that our approach outperforms those based on adversarial training.

\textbf{Lottery Ticket Hypothesis.} This hypothesis, introduced in~\cite{frankle2018the}, states that overparameterized neural networks contain sparse subnetworks that can be trained in isolation to achieve competitive performance.
These subnetworks are called the \textit{winning tickets}.
Based on this interesting observation, \cite{zhou2019deconstructing, ramanujan2020s} further proposed the \textit{Strong Lottery Ticket Hypothesis}.
They showed that there exist winning tickets with competitive performance even without training.
Furthermore, \cite{chijiwa2021pruning} proposed an iterative randomization scheme to reduce the size of the network in which one searches for the winning tickets.
\zzhaocamera{\cite{diffenderfer2021multiprize} introduced the \textit{Multi-Prize Lottery Ticket Hypothesis} to learn compact yet accurate binary networks by pruning and quantizing randomly weighted DNNs. \cite{diffenderfer2021a} showed that ``lottery-ticket style'' approaches can also improve robustness against corruption in the frequency domain.}

\change{The recent work of~\cite{fu2021drawing} combines robustness with the \textit{Strong Lottery Ticket Hypothesis} and demonstrates the existence of robust sub-networks within a random network. Here, we focus on lighter-weight binary networks, and introduce an adaptive pruning strategy and last batch normalization layer to achieve higher pruning rates than~\cite{fu2021drawing} while maintaining a competitive accuracy.
}

\section{Methodology} \label{sec:method}

\subsection{\change{Preliminaries: \textit{Edge-Popup} under Adversarial Attacks}} \label{subsec:edgepopup}

\change{Let us first formulate the problem in a similar manner to~\cite{fu2021drawing}.}
We consider a neural network $f$ parameterized by $\rvw \in \sR^n$. For an input sample $(\rvx, y)$, the neural network outputs $f(\rvw, \rvx)$.
$\gL(f(\rvw, \rvx), y)$ then represents the training loss objective, where $\gL$ is the softmax cross-entropy loss.
Given a dataset $\{(\rvx_i, y_i)\}_{i = 1}^N$, an adversarial budget $\gS_\epsilon$, and a predefined pruning rate $r$, we search for a binary pruning mask $\rvm$ that solves the following optimization problem:
\begin{equation}
\begin{aligned}
\min_\rvm \frac{1}{N} \sum_{i = 1}^N \max_{\Delta_i \in \gS_\epsilon} \gL\left(f(\rvw \odot \rvm, \rvx_i + \Delta_i), y_i\right) \
s.t.\ \rvm \in \{0, 1\}^n, sum(\rvm) = (1 - r)n.
\end{aligned} \label{eq:problem}
\end{equation}

Here, function $sum$ calculates the summation of all the elements in a vector.
In contrast to adversarial training, we do not optimize the model parameters $\rvw$ in (\ref{eq:problem}); instead $\rvw$ contains randomly-initialized parameters that are kept fixed during optimization.
As such, our algorithm aims to find a pruned network structure, encoded via the $n$-dimensional binary vector $\rvm$, corresponding to a robust subnetwork.
Since the mask $\rvm$ is a discrete vector, it cannot be directly optimized by gradient-based methods. To overcome this, we replace it with a continuous ``score'' variable, $\rvs \in \sR^n$, from which we  calculate the mask as
\begin{equation}
\begin{aligned}
\rvm = M(\rvs, r)\;,
\end{aligned} \label{eq:mask}
\end{equation}
where $M$ is a binarization function. It constructs a binary mask from the continuous-valued scores based on a pruning strategy and a required pruning rate $r$.
\change{The pruning strategy can be global or layer-wise, and retains the parameters with the highest scores.
In the layer-wise case, it automatically determines the number of parameters retained in each layer.}

\change{To update the scores $\rvs$, we use the same \textit{edge-popup} strategy as in~\cite{fu2021drawing, raghunathan2018certified}.
Specifically, we use straight through estimation~\cite{bengio2013estimating} to calculate the gradient $\partial \gL / \partial \rvs$.
Note that the approximation made in straight through estimation does not affect the adversarial example generation in adversarial training.
Our experiments will show that we can effectively generate adversarial examples by PGD.
We provide the pseudo-code of our algorithm in Appendix~\ref{sec:app_algorithm}.
}


\subsection{Adaptive Pruning} \label{subsec:adaptivepruning}

\change{
As mentioned above, in addition to the given pruning rate $r$, the binarization function $M$ in Equation~(\ref{eq:mask}) also depends on the pruning strategy.
\zzhao{\cite{gui2019model}} uses \textit{global pruning}, retaining the $(1 - r)n$ parameters with the highest scores, regardless of which layer they belong to.
However, global pruning does not consider the topology of the network, and the fact that the magnitude of the scores $\rvs$ can differ from layer to layer.
Furthermore, when the pruning rate $r$ is close to $1$, global pruning may prune some layers entirely, thus causing a trivial performance.
Therefore, other works~\zzhao{\cite{fu2021drawing, ramanujan2020s, wang2020hydra}} use layer-wise pruning strategies.
For an $L$-layer network with $\{n_i\}_{i = 1}^L$ parameters and a predefined pruning rate $r$, such strategies first allocate the number of parameters $\{m_i\}_{i = 1}^L$ to retain in each layer, and then retain the parameters with the highest scores in each layer.

In Appendix~\ref{subsec:prune_strategy}, we discuss two special cases of layer-wise pruning: \textit{fixed pruning rate} and \textit{fixed number of parameters}.
With \textit{fixed pruning rate}, we have $1 - r = \frac{m_1}{n_1} = \frac{m_2}{n_2} = ... = \frac{m_L}{n_L}$.
Theorem~\ref{thm:fix_pr} indicates that this maximizes the size of the search space of the subnetwork.
However, this strategy might retain too few parameters for the small layers when $r$ is big, which has two serious drawbacks: 1) It greatly limits the expression power of the network; 2) it makes the \textit{edge-popup} algorithm less stable, because adding or removing a single parameter then has a large impact on the network's output.
This instability becomes even more pronounced in the presence of adversarial samples, because the gradients of the model parameters are more scattered than when training on clean inputs~\cite{liu2020loss}.

With \textit{fixed number of parameters}, we have $m_1 = m_2 = ... = m_L$.
When the allocated number of retained parameters exceeds the total number of the original parameters in one layer, we leave this layer totally unpruned.
As shown in Theorem~\ref{thm:maxpath}, this strategy maximizes the number of paths from the input layer to the output layer.
In contrast to the \textit{fixed pruning rate}, a \textit{fixed number of parameters} may retain too many parameters in small layers.
In the extreme case, some layers may be entirely unpruned when the pruning rate $r$ is small.
This is problematic in our settings, since the model parameters are random and not updated.
}

In other words, the two strategies discussed above are two extremes: the \textit{fixed pruning rate} one suffers when $r$ is big, whereas the \textit{fixed number of parameters} one suffers when $r$ is small.
To address this, we propose a strategy in-between these two extremes.
Specifically, we determine the number of parameters retained in each layer by solving the following system of equations:
\begin{equation}
\begin{aligned} \label{eq:strategy}
1 - r = \frac{\sum_{i = 1}^L m_i}{\sum_{i = 1}^L n_i}, \frac{m_1}{n_1^p} = \frac{m_2}{n_2^p} = ... = \frac{m_L}{n_L^p}\;,
\end{aligned}
\end{equation}
where $p \in [0, 1]$ is a hyper-parameter controlling the trade-off between the two extreme cases.
When $p = 0$, the strategy~(\ref{eq:strategy}) is the \textit{fixed number of parameters} one. 
When $p = 1$, the strategy becomes the \textit{fixed pruning rate} one.
By setting $0 < p < 1$, we can retain a higher proportion of parameters in the smaller layers without sacrificing the big layers too much.
We call this strategy \textit{adaptive pruning}.
As discussed above, the strategy obtained with $p = 1$ tends to fail with a big $r$, while the strategy resulting from setting $p = 0$ tends to fail with a small $r$.
This indicates that we need to assign small values of $p$ given a big $r$ and big values of $p$ otherwise.
We validate this and study the influence of $p$ on the results of our approach in experiments.

\subsection{Binary Initialization and Last Normalization Layer} \label{subsec:binary}

\change{
Our work focuses on binary networks, which have a much smaller memory footprint than full-precision networks but are more challenging to train.
To address this, we therefore study the influence of the binary initialization scheme and introduce a \textit{last normalization layer} approach to facilitate training and boost the performance.
}


\textbf{Binary initialization.}
The empirical studies of~\cite{ramanujan2020s} demonstrate the importance of the initialization scheme on the performance of a pruned network.
As a result, \cite{ramanujan2020s} proposes the \textit{Signed Kaiming Constant} initialization: The parameters in layer $i$ are uniformly sampled from the set $\left\{-\sqrt{\frac{2}{l_{i - 1}(1 - r)}}, \sqrt{\frac{2}{l_{i - 1}(1 - r)}}\right\}$, where $l_{i - 1}$ represents the fan-out of the previous layer. 
Correspondingly, the scores $\rvs$ are initialized based on a uniform distribution $U[-\sqrt{\frac{1}{l_{i - 1}}}, \sqrt{\frac{1}{l_{i - 1}}}]$.

The magnitude of the \textit{Signed Kaiming Constant} initialization is carefully calculated to keep the variance of the intermediate activations stable from the input to the output.
In modern deep neural networks, the convolutional layers, potentially together with activation functions, are typically followed by a batch normalization layer.
In~\cite{ramanujan2020s} and our settings, these batch normalization layers only estimate the running statistics of their inputs, they do not have trainable parameters representing affine transformations.
Because of these batch normalization layers, the magnitudes of the convolutional layers do not affect the outputs of the ``convolution-batch norm'' blocks.
Furthermore, the fully-connected layers on top of the convolutional ones are homogeneous\footnote{We call a function $f$ homogeneous if it satisfies $\forall x\ \forall a \in \sR^+$, $f(ax) = af(x)$.} because their bias terms are always initialized to zero and not updated during training.
The activation functions we use, such as ReLU or leaky ReLU~\cite{maas2013rectifier}, are also homogeneous.
Therefore, the magnitudes of parameters in these fully-connected layers do not change the predicted labels of the model either.

Based on the analysis above, we conclude that the magnitudes of the model parameters at initialization do not change the predicted labels.
Therefore, we propose to scale the model parameters $\rvw$ in all linear layers, i.e., convolutional and fully-connected ones, so that they are all sampled from $\{-1, +1\}$.
Correspondingly, the scores $\rvs$ are initialized based on a uniform distribution $[-a, a]$, where $a$ is a factor controlling the variance.

Our binary initialization scheme is beneficial to model compression and acceleration, since there are no longer multiplication operations in linear layers.
We discuss the efficiency improvement of the binary networks in detail in Appendix~\ref{subsec:binary_analysis}.
\change{Theoretically, for the RN34 models we use in this paper, binary initialization can save approximately $45\%$ and $32\%$ FLOP operations compared with their full precision counterparts in the training phase and evaluation phase, respectively.
Since we use irregular pruning in our method, taking full advantage of this improvement requires lower-level and hardware customization.}

\textbf{Last normalization layer.}
Although scaling the model parameters does not affect the expression power of the network nor change the predicted label given the input, it does change the optimization landscape of the problem (\ref{eq:problem}), because the softmax cross-entropy function used to calculate the loss objective is not homogeneous.
Compared with the \textit{Signed Kaiming Constant} method, our binary initialization multiplies the parameters initialized in the last layer by $\sqrt{\frac{l_{L-1} (1 - r)}{2}}$.
Therefore, the output logits fed to the softmax cross-entropy function are also multiplied by the same factor.
In practice, $\sqrt{\frac{l_{L-1} (1-r)}{2}} \gg 1$ greatly increases the output logits.
Large logits will cause numerical instability and thus greatly worsen the optimization performance.
In particular, our detailed analysis in Appendix~\ref{subsec:normalization} shows that such scaling causes gradient vanishing for correctly classified inputs and, even worse, gradient exploding for misclassified ones.

To address this issue, we add another 1-dimensional batch normalization layer at the end of the model, just before the softmax layer.
The analysis in Appendix~\ref{subsec:normalization} shows this normalization layer cancels out the multiplication factor applied to the weights in the last layer and thus facilitates the optimization.
In our experiments, we show that this normalization layer greatly improves the performance of both the \textit{Signed Kaiming Constant} method and our \textit{Binary Initialization} one.
Furthermore, the last normalization layer also makes the performance more robust to different score $\rvs$ initializations.

\section{Experiments}

In this section, we present extensive experimental results to validate our approach.
First, we describe an ablation study and sensitivity analysis.
Then, we compare our performance with existing works, which achieve robustness and compression in either full-precision or binary cases.
We also include adversarial training~\cite{madry2017towards} as a baseline.
Finally, we analyze the structure of the pruned networks that we obtain.
We show some interesting patterns of these post-pruning networks, suggesting the potential of our approach for more effective compression.

Unless explicitly stated otherwise, we use a 34-layer Residual Network (RN34)~\cite{he2016deep}, the same as the one in~\cite{ramanujan2020s, wang2020hydra}.\footnote{Note that the RN34 used in these papers and ours differs from the WideRN34-10 used in~\cite{madry2017towards, wu2020adversarial}, which is larger and has almost twice the number of trainable parameters.}
We use the CIFAR10 dataset~\cite{krizhevsky2009learning} in the ablation study; we also use the CIFAR100 dataset~\cite{krizhevsky2009learning} \zzhao{and the ImageNet100 dataset~\cite{deng2009imagenet, douillardlesort2021continuum}} in the comparisons with the baselines. \footnote{All these datasets are free for non-commerical use. CIFAR10 and CIFAR100 are downloadable on PyTorch. \zzhaocamera{ImageNet can be downloaded from~\href{https://www.kaggle.com/competitions/imagenet-object-localization-challenge/data}{Kaggle} and the subset we use can be found in~\href{https://github.com/Continvvm/continuum/blob/838ad2ba3571f1563627301c30152c0f07d3cffa/continuum/datasets/imagenet.py}{Contunuum's documentation}.}}
\zzhao{We train the models for $400$ epochs on CIFAR10/100 and $100$ epochs on ImageNet100. We use a cosine annealing learning rate scheduler with an initial value of $0.1$.}
Unless specified, we employ PGD attacks~\cite{madry2017towards} to generate adversarial examples during training, but we use AutoAttack (AA)~\cite{croce2020reliable} for our robustness evaluation. 
While PGD is much faster than AutoAttack and thus suitable for training, AutoAttack is the current state-of-the-art attack method, and we thus consider it a more reliable metric of robustness.
We use an $l_\infty$ norm-based adversarial budget, and the perturbation strength $\epsilon$ is $8 / 255$ for CIFAR10, $4 / 255$ for CIFAR100 \zzhao{and $2 / 255$ for ImageNet100.}
More details about the experimental settings and hyper-parameters are listed in Appendix~\ref{subsec:exp_setting}.

\subsection{Ablation Study and Sensitivity Analysis}


\textbf{Pruning Strategy and Pruning Rates.} We first focus on binary initialization and on the models with the last batch normalization layer (LBN).
We compare the performance of our method under different pruning rates $r$ and \textit{adaptive pruning strategies} with different values of $p$.
The scores $\rvs$ are initialized from a uniform distribution $U[-0.01, 0.01]$.

Our results are summarized in Table~\ref{tbl:sp_ratio}, in which we include $7$ different values of pruning rate $r$ and $7$ different values of $p$ in the \textit{adaptive pruning strategy}.
First, we notice that the best performance is achieved when $r = 0.99$ and $p = 0.1$.
For the \textit{fixed pruning rate} strategy ($p = 1.0$), the best performance is achieved when $r = 0.8$.
Compared with the vanilla (i.e., non-adversarial) case in~\cite{ramanujan2020s}, which uses the \textit{fixed pruning rate} strategy and shows that $r = 0.5$ achieves the best clean accuracy, the best performance for robust accuracy is achieved at a much higher pruning rate.
This interesting observation is also consistent with the existing work~\cite{croce2019provable}, which shows that adversarial training implicitly encourages sparse convolutional kernels.

\begin{table*}[!ht]
\small
\centering
\begin{tabular}{p{2.cm}<{\centering}p{1.1cm}<{\centering}p{1.1cm}<{\centering}p{1.1cm}<{\centering}p{1.2cm}<{\centering}p{1.2cm}<{\centering}p{1.4cm}<{\centering}p{1.4cm}<{\centering}}
\Xhline{4\arrayrulewidth}
Prune Strategy & $r = 0.5$ & $r = 0.8$ & $r = 0.9$ & $r = 0.95$ & $r = 0.99$ & $r = 0.995$ & $r = 0.998$ \\
\hline
$p = 0.0$ &  2.16 &  6.86 & 23.01 & 41.61 & 44.60 & 40.70 & \textbf{34.97} \\
$p = 0.1$ &  4.35 & 15.03 & 28.12 & 42.65 & \textbf{44.88} & \textbf{40.97} & 33.09 \\
$p = 0.2$ &  8.01 & 19.21 & 27.99 & 43.72 & 42.92 & 40.52 & 32.99 \\
$p = 0.5$ &  9.21 & 32.70 & 42.84 & 43.62 & 42.45 & 40.55 & 30.08 \\
$p = 0.8$ & 28.90 & 41.51 & \textbf{43.64} & \textbf{43.88} & 39.12 & 33.61 & 28.07 \\
$p = 0.9$ & 39.09 & 41.71 & 43.07 & 42.28 & 38.68 & 33.89 & 17.43 \\
$p = 1.0$ & \textbf{42.85} & \textbf{43.23} & 42.13 & 41.12 & 34.57 & 26.67 & 20.56 \\
\Xhline{4\arrayrulewidth}
\end{tabular}
\caption{Robust accuracy (in \%) on the CIFAR10 test set under different pruning rates $r$ and values of $p$ in \textit{adaptive pruning}. The best result for each pruning rate is marked in bold.} \label{tbl:sp_ratio}
\end{table*}

Table~\ref{tbl:sp_ratio} further demonstrates the benefits of our \textit{adaptive pruning strategy}.
For larger pruning rates $r$, a smaller value of $p$ prevails; for smaller pruning rates, a bigger value of $p$ prevails.
This is consistent with our analysis in Section~\ref{subsec:adaptivepruning}.
In particular, compared with the best results for a fixed pruning rate strategy ($p = 1.0$, $r = 0.8$), which is the pruning strategy used in~\cite{ramanujan2020s}, our best adaptive pruning ($p = 0.1$, $r = 0.99$) achieves not only better performance but also a higher pruning rate.
That is to say, using our \textit{adaptive pruning strategy} improves both robustness and compression rates.

In Figure~\ref{fig:stability} of Appendix~\ref{subsec:p_sensitivity_analysis}, we provide the learning curves when $r = 0.99$ and when $r = 0.5$.
Regardless of the pruning rate $r$, these curves indicate the importance of the pruning strategy: a well chosen $p$ value not only improves the performance but also makes training more stable.

\textbf{Last Normalization Layer.} We then study how the last batch normalization layer (LBN) introduced in Section~\ref{subsec:binary} affects the performance.
We focus on the \textit{binary initialization} first and report the performance of models with and without the last normalization layer under different values of $a$, the hyper-parameter controlling the variance of the initial score $\rvs$.
Based on the results in Table~\ref{tbl:sp_ratio}, we use the \textit{adaptive pruning strategy} with $p = 0.1$ and a pruning rate $r = 0.99$.


\begin{table}[!htb]
\begin{minipage}{.4\textwidth}
\centering
\small
\begin{tabular}{p{1.8cm}<{\centering}p{1.1cm}<{\centering}p{1.1cm}<{\centering}}
\Xhline{4\arrayrulewidth}
Value of $a$ in & \multirow{3}{*}{no LBN} & \multirow{3}{*}{LBN} \\
in Score & & \\
Initialization & & \\
\hline
0.001 & 33.08 & \textbf{45.06} \\
0.01  & 39.96 & 44.88 \\
0.1   & \textbf{41.01} & 44.63 \\
1     & 31.04 & 44.41 \\
\Xhline{4\arrayrulewidth}
\end{tabular}
\vspace{0.5cm}
\caption{Robust accuracy (in \%) on the CIFAR10 test set for models with and without the last batch normalization layer (LBN) under different values of $a$ for score $\rvs$ initialization. The best results are marked in bold.}
\label{tbl:bn_ablation}
\end{minipage}
\begin{minipage}{.05\textwidth}
~~~
\end{minipage}
\begin{minipage}{.55\textwidth}
\centering
\small
\begin{tabular}{p{1.2cm}<{\centering}p{1.1cm}<{\centering}p{1.1cm}<{\centering}p{1.1cm}<{\centering}p{1.1cm}<{\centering}}
\Xhline{4\arrayrulewidth}
Prune & \multicolumn{2}{c}{Signed KC} & \multicolumn{2}{c}{Binary} \\
Strategy & no LBN & LBN & no LBN & LBN \\
\hline
$p = 0.0$ & 39.38 & 42.83 & 40.94 & 44.65 \\
$p = 0.1$ & 39.62 & 45.01 & \textbf{41.01} & \textbf{45.06} \\
$p = 0.2$ & 36.66 & \textbf{45.04} & 37.85 & 41.58 \\
$p = 0.5$ & \textbf{39.98} & 42.64 & 40.61 & 39.95 \\
$p = 0.8$ & 37.96 & 41.71 & 35.15 & 38.95 \\
$p = 0.9$ & 34.75 & 40.14 & 35.64 & 35.81 \\
$p = 1.0$ & 36.88 & 39.32 & 30.02 & 30.62 \\
\Xhline{4\arrayrulewidth}
\end{tabular}
\vspace{0.2cm}
\caption{Robust accuracy (in \%) on the CIFAR10 test set with the \textit{Signed Kaiming Constant} (Signed KC) and the binary initialization. We include models both with and without the last batch normalization layer (LBN). The best results are marked in bold.} \label{tbl:main_ablation}
\end{minipage}
\end{table}


The results are provided in Table~\ref{tbl:bn_ablation} and clearly show that the last batch normalization layer (LBN) greatly improves the performance.
Furthermore, LBN makes the performance much less sensitive to the initialization of the scores, which in practice facilities the hyper-parameter selection.

\textbf{Initialization Scheme.} Finally, we compare the performance of the \textit{binary initialization} with the \textit{Signed Kaiming Constant}.
We fix the pruning rate to $r = 0.99$ and employ an adaptive pruning strategy with different values of $p$.
Our results are summarized in Table~\ref{tbl:main_ablation}.
For binary initialization, we use the optimal initialization scheme of the score $\rvs$ from Table~\ref{tbl:bn_ablation}; for \textit{Signed Kaiming Constant} initialization, we use the optimal setting from~\cite{ramanujan2020s} to initialize $\rvs$.


Based on the results in Table~\ref{tbl:main_ablation}, we can conclude that the \textit{binary initialization} achieves a comparable performance with the \textit{Signed Kaiming Constant}.
Furthermore, the last batch normalization layer also improves the performance when using the \textit{Signed Kaiming Constant}.
We show in Table~\ref{tbl:vanilla_ablation} of Appendix~\ref{subsec:vanilla} that these conclusions are also valid in a non-adversarial setting.

\subsection{Comparison with Existing Methods} \label{subsec:compare_existing}

\textbf{Baselines.} In this section, we compare our approach with the state-of-the-art methods targeting model compression and robustness.
\change{Specifically, we include FlyingBird, FlyingBird+\cite{chen2022sparsity}, Bayesian Connectivity Sampling (BCS)~\cite{ozdenizci2021training}, Robust Scratch Ticket (RST)~\cite{fu2021drawing}, HYDRA~\cite{wang2020hydra} and ATMC~\cite{gui2019model}, as well as adversarial training (AT)~\cite{madry2017towards} with early stopping~\cite{rice2020overfitting}.}
Given our previous results, we fix the pruning rate to $r = 0.99$.
For adversarial training, we use the full RN34 model and some smaller networks with approximately the same number of parameters as our pruned models.
These smaller networks have the same architecture as the RN34 except that they have fewer channels.
The details of these small networks are shown in Table~\ref{tbl:smallnet} of Appendix~\ref{subsec:exp_setting}.
We follow the official implementations of all the baselines, and thus, unlike in our method, the normalization layers in all the baselines that update model parameters have an affine transformation with trainable parameters.

ATMC supports quantization but its parameterization introduces learnable quantized values.
That is, \cliucamera{although} the models obtained by ATMC's 1-bit quantization have only two parameter values in each layer, these values are different from layer to layer and are not necessarily $-1$ and $+1$.
This means that, compared with the binary networks obtained with our method, those from ATMC have more \cliucamera{trainable parameters and thus} flexibility.
Nevertheless, we still include ATMC for comparison in the case of binary networks.
Similarly to our method, RST does not update the model parameters.
It initializes the model parameters with full-precision values, and we thus only provide full-precision results for RST.
The other baselines and AT are not designed for quantization and do not inherently support binary networks.
To address this, we use \textit{BinaryConnect}~\cite{courbariaux2015binaryconnect} to replace the model's linear layers so that their parameters are binary.
\textit{BinaryConnect} generates binarized model parameters by taking the sign of the weights during the forward pass, and uses straight-through estimation~\cite{bengio2013estimating} for gradient calculation.

Our method uses \textit{binary initialization} and the last batch normalization layer, so the models we obtained are inherently binary.
In addition to using PGD-based adversarial examples, we accelerate our method by using adversarial examples based on FGSM~\cite{goodfellow2014explaining} with ATTA~\cite{zheng2020efficient}.
FGSM with ATTA generates adversarial examples by one-step attacks with accumulated perturbations across epochs. This is much cheaper than the $10$-step PGD attacks.
\cliucamera{For CIFAR10 and CIFAR100, we provide the results of our method when using FGSM with ATTA as ``Ours(fast)'' in Table~\ref{tbl:main comparison} for comparison.
For ImageNet100, since the dataset is bigger and the images are of much higher resolution, the computational cost for multi-step PGD is huge.
Therefore, we use FGSM with ATTA to generate adversarial examples for all methods on ImageNet100.
To decrease the memory overhead introduced by ATTA, we only store the downsampled perturbations in the current epoch for the perturbation initialization of the next epoch.
We provide the pseudo-code and more details in Appendix~\ref{sec:app_algorithm}.}

\begin{table*}[!t]
\centering
\begin{tabular}{p{1.7cm}p{2.1cm}<{\centering}p{1.6cm}<{\centering}p{0.8cm}<{\centering}p{0.8cm}<{\centering}p{0.8cm}<{\centering}p{0.8cm}<{\centering}p{0.8cm}<{\centering}p{0.8cm}<{\centering}}
\Xhline{4\arrayrulewidth}
\multirow{2}{*}{Method} & \multirow{2}{*}{Architecture} & Pruning & \multicolumn{2}{c}{CIFAR10} & \multicolumn{2}{c}{CIFAR100} & \multicolumn{2}{c}{ImageNet100} \\
& & Strategy & FP & Binary & FP & Binary & FP & Binary \\
\hline
AT & RN34       & Not Pruned & 43.26 & 40.34 & 36.63 & 26.49 & 53.92 & 34.20 \\
AT & RN34-LBN   & Not Pruned & 42.39 & 39.58 & 35.15 & 32.98 & 55.14 & 35.36 \\
\hdashline
AT & Small RN34 & Not Pruned & 38.81 & 26.03 & 27.68 & 15.85 & 25.40 & 10.44 \\
FlyingBird  & RN34 & Dynamic & \underline{45.86} & 34.37 & \underline{35.91} & 23.32 & 37.70 & 9.54 \\
FlyingBird+ & RN34 & Dynamic & 44.57 & 33.33 & 34.30 & 22.64 & 37.70 & 9.52 \\
BCS         & RN34 & Dynamic & 43.51 & - & 31.85 & - & - & - \\
RST & RN34     & $p = 1.0$ & 34.95 & - & 21.96 & - & 17.54 & - \\
RST & RN34-LBN & $p = 1.0$ & 37.23 & - & 23.14 & - & 15.36 & - \\
HYDRA & RN34   & $p = 0.1$ & 42.73 & 29.28 & 33.00 & 23.60 & \underline{43.18} & 18.22 \\
ATMC  & RN34   & Global    & 34.14 & 25.62 & 25.10 & 11.09 & 22.18 & 5.78 \\
ATMC  & RN34   & $p = 0.1$ & 34.58 & 24.62 & 25.37 & 11.04 & 23.52 & 4.58 \\
Ours       & RN34-LBN & $p = 0.1$ & - & \textbf{45.06} & - & \textbf{34.83} & \multirow{2}{*}{-} & \multirow{2}{*}{\textbf{33.04}} \\
Ours(fast) & RN34-LBN & $p = 0.1$ & - & 40.77 & - & 34.45 & & \\
\Xhline{4\arrayrulewidth}
\end{tabular}
\caption{Robust accuracy (in \%) on the CIFAR10, CIFAR100 \cliucamera{and ImageNet100} test sets for the baselines and our proposed method. ``RN34-LBN'' represents ResNet34 with the last batch normalization layer. ``Small RN34'' refers to Small RN34-p0.1 in Table~\ref{tbl:smallnet} of Appendix~\ref{subsec:exp_setting}. The pruning rate is set to $0.99$ except for the not-pruned methods. \cliucamera{Among the pruned models}, the best results for the full-precision (FP) models are underlined; the best results for the binary models are marked in bold. \cliucamera{The values of $\epsilon$  for CIFAR10, CIFAR100 and ImageNet100 are $8 / 255$, $4 / 255$ and $2 / 255$, respectively. ``-'' means not applicable or trivial performance.}}~\label{tbl:main comparison}
\end{table*}
\textbf{Results.} 
Our main results on CIFAR10, CIFAR100 \cliucamera{and ImageNet100} are summarized in Table~\ref{tbl:main comparison}, where we report the robust accuracy under AutoAttack (AA), \cliucamera{which is considered as a reliable evaluation metric for robustness~\cite{croce2020reliable}.}
The results of all baselines are based on their default settings in architecture and pruning strategy based on publicly available codes.
\footnote{Publicly available code on GitHub: FlyingBird/FlyingBird+: \href{https://github.com/VITA-Group/Sparsity-Win-Robust-Generalization}{VITA-Group/Sparsity-Win-Robust-Generalization}; BCS: \href{https://github.com/IGITUGraz/SparseAdversarialTraining}{IGITUGraz/SparseAdversarialTraining}; RST:  \href{https://github.com/RICE-EIC/Robust-Scratch-Ticket}{RICE-EIC/Robust-Scratch-Ticket}; HYDRA: \href{https://github.com/inspire-group/hydra}{inspire-group/hydra}; ATMC: \href{https://github.com/VITA-Group/ATMC}{VITA-Group/ATMC}. All the codes are free to use for non-commerical purposes.}
The exceptions are that we also include adaptive pruning ($p = 0.1$) for HYDRA, ATMC, and the last batch normalization layer for RST, because we noticed such changes to improve their performance.

Our method using the \textit{adaptive pruning} strategy ($p = 0.1$) achieves better performance than all baselines in case of binary models.
\cliucamera{On CIFAR10 and CIFAR100, we also achieve comparable performance to methods using full-precision models.}
Furthermore, our method achieves results comparable with AT on the original unpruned models that has $100 \times$ more trainable parameters.
In addition, our method based on \cliucamera{FGSM with ATTA, which is much faster than multi-step PGD,} also achieves better performance than all baselines in the case of binary networks.
\cliucamera{On ImageNet100, our method, which aims to train binary networks, also outperforms most full-precision networks trained by the baselines.
BCS yields almost trivial performance on ImageNet100 (< 3\%) and is thus not included.
This suggests that BCS cannot converge using a high compression rate and facing a complicated dataset.
Compared with adversarial training on the full network, which has $100$ times as many parameters as ours, we achieve comparable performance with the binary networks, but worse performance than the full-precision networks.
Note that fitting the high-dimensional ImageNet100 dataset under adversarial attacks using only 1\% of the binary parameters is extremely challenging.
As demonstrated in Table~\ref{tbl:main comparison}, many baselines only achieve low robust accuracy in this setting.}

For all baselines except RST, the last normalization layer does not improve the performance; it even hurts the performance in the full-precision cases.
This is because these baselines (except RST) update the model parameters $\rvw$.
In the full precision cases, the magnitude of $\rvw$, and thus of the output logits, is automatically adjusted during training.
The issue resulting from large output logits that we pointed out in Section~\ref{sec:method} does thus not happen in these cases, so the last batch normalization layer is not necessary.
In practice, we observed this layer to slow down the training convergence of these models.

For the pruning strategy, the proposed \textit{adaptive pruning} strategy ($p = 0.1$) consistently achieves better performance than the \textit{fix pruning rate} strategy ($p = 1.0$) and than \textit{global pruning}.
FlyingBird, FlyingBird+ and BCS dynamically assign retrained parameters during training, which has similar benefits to adaptive pruning but at the cost of training efficiency~\cite{chen2022sparsity}.
\cliucamera{Furthermore, although the value of $p$ is selected based on the ablation study on CIFAR10, it also performs well on CIFAR100 and ImageNet100.
This observation indicates that for a fixed value of $r$, the selection of $p$ generalizes well across different datasets.}

We further compare the baseline methods with various settings such as adding the last batch normalization layer, changing the pruning strategy, using different AT methods, and provide a complete set of comparison results in Appendix~\ref{subsec: more baseline results}.
\cliucamera{The conclusions drawn from Table~\ref{tbl:main comparison} remain valid.}

\begin{table}[!ht]
\centering
\begin{tabular}{p{1.9cm}p{2.5cm}<{\centering}p{1.8cm}<{\centering}p{0.8cm}<{\centering}p{0.8cm}<{\centering}p{0.8cm}<{\centering}p{0.8cm}<{\centering}}
\Xhline{4\arrayrulewidth}
\multirow{2}{*}{Method} & \multirow{2}{*}{Architecture} & Pruning & \multicolumn{2}{c}{RN18} & \multicolumn{2}{c}{RN50} \\
& & Strategy & FP & Binary & FP & Binary \\
\hline
AT    & RN & Not Pruned & 41.50 & 39.13 & 43.24 & \zzhaocamera{31.18} \\ 
AT    & RN-LBN & Not Pruned & 42.25 & 39.86 & 44.33 & 37.25 \\
\hdashline
AT & Small RN & Not Pruned & 28.13 & 30.35 & 26.03 & \zzhaocamera{32.25} \\ 
FlyingBird & RN & Dynamic & \underline{42.15} & 27.08 & 35.91 & 26.33 \\
FlyingBird+ & RN & Dynamic & 38.55 & 27.84 & 29.54 & 25.40 \\
BCS & RN & Dynamic & 39.60 & 21.46 & 41.85 & 17.54 \\
RST & RN & $p=1.0$ & 31.98 & - & 35.40 & - \\
RST & RN-LBN & $p=1.0$ & 33.27 & - & 34.71 & - \\
HYDRA & RN & $p = 0.1$  & 40.20 & 30.90 & \underline{44.14} & 22.36 \\
ATMC  & RN & Global & 32.21 & 17.73 & 25.23 & 6.82 \\
ATMC  & RN & $p = 0.1$ & 32.31 & 19.67 & 33.61 & 16.12 \\
Ours & RN-LBN & $p = 0.1$ & - & \textbf{39.65} & - & \textbf{42.72} \\
Ours (fast) & RN-LBN & $p = 0.1$ & - & 30.86 & - & 37.93 \\
\Xhline{4\arrayrulewidth}
\end{tabular}
\caption{\cliucamera{Robust accuracy (in \%) on the CIFAR10 test set for AT, FlyingBird(+), BCS, RST, HYDRA, ATMC and our proposed method on the RN18 and RN50 models. ``RN-LBN'' represents networks with the last batch normalization layer. Among the compressed models, the best results for full precision (FP) models are underlined; the best results for binary models are marked in bold.}} \label{tbl:comparison rn18 rn50}
\end{table}

All the results in Table~\ref{tbl:main comparison} are based on a RN34 architecture, \cliucamera{Table~\ref{tbl:comparison rn18 rn50} provides the results on CIFAR10 using a smaller 18-layer network (RN18) and a larger 50-layer network (RN50).
These results confirm the effectiveness of our method on different network architectures.}

\cliucamera{In addition to robust accuracy, Table~\ref{tbl:comparison vanilla} in Appendix~\ref{subsec:app_vanilla_accuracy} demonstrates the accuracy on the clean test set for the models in Table~\ref{tbl:main comparison}.}
Our method also yields competitive performance on clean inputs.
Specifically, we achieve the best performance among all methods for binary networks.
Combining the results in Table~\ref{tbl:main comparison} and~\ref{tbl:comparison vanilla}, we conclude that our method yields a better trade-off between accuracy on clean inputs and accuracy on adversarially perturbed inputs.

\cliucamera{Finally, vanilla training can be considered as a special case of adversarial training, where $\epsilon = 0$.
Therefore, our method, as well as all baselines, are applicable to vanilla training.
The results when $\epsilon = 0$ are provided in Table~\ref{tbl:vanilla} of Appendix~\ref{subsec:nonadversarial}.
Our method achieves the best performance among the pruned binary networks.
This indicates that our method is competitive under difference adversarial budgets.
}

\subsection{Analysis of the Subnetwork Patterns}
\label{subsec:main subnetwork pattern}

In this work, we use irregular pruning.
Compared with regular pruning, irregular pruning is more flexible but less structured, which means that it requires lower-level customization to fully take advantage of parameter sparsity for acceleration.
However, visualizing the masks $\rvm$ of the convolutional layers in our pruned binary network with a pruning rate $r = 0.99$ allowed us to find that the mask is structured to some degree.
For example, we visualize the mask of a convolutional layer with $256$ input channels and $256$ output channels in Figure~\ref{fig:mask_full} of Appendix~\ref{subsec:mask}.
We notice that the retained parameters are quite concentrated \cliucamera{and structured}: \cliucamera{Most} retained parameters concentrate on few input or output channels, while many other channels ($40\%$ of the total) are completely pruned.

\begin{figure}[!ht]
\centering
\begin{subfigure}{0.23\textwidth}
    \centering
    \includegraphics[width=\linewidth]{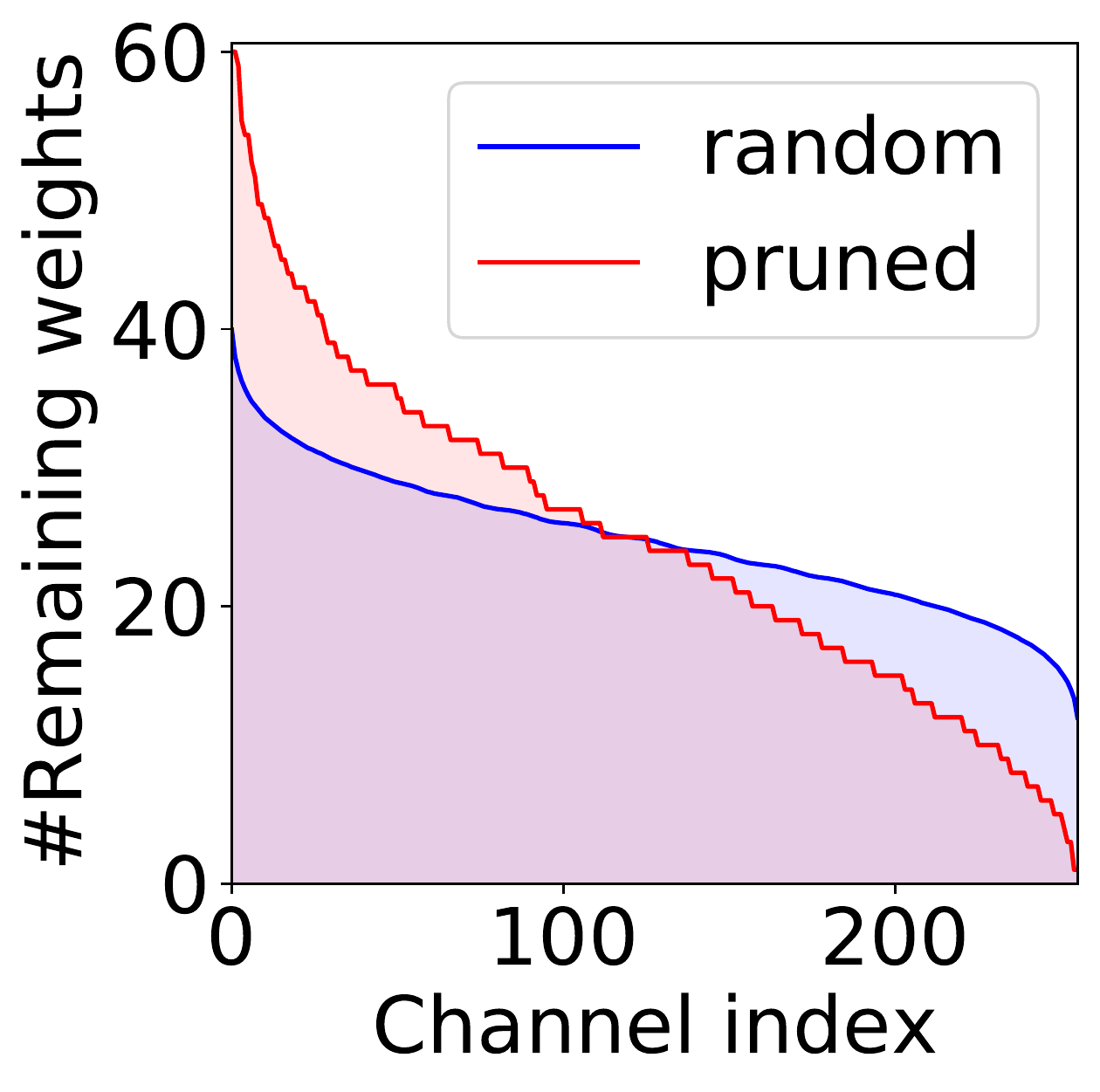}
    \caption{L1 Input Channel}
\end{subfigure}
~~
\begin{subfigure}{0.23\textwidth}
    \centering
    \includegraphics[width=\linewidth]{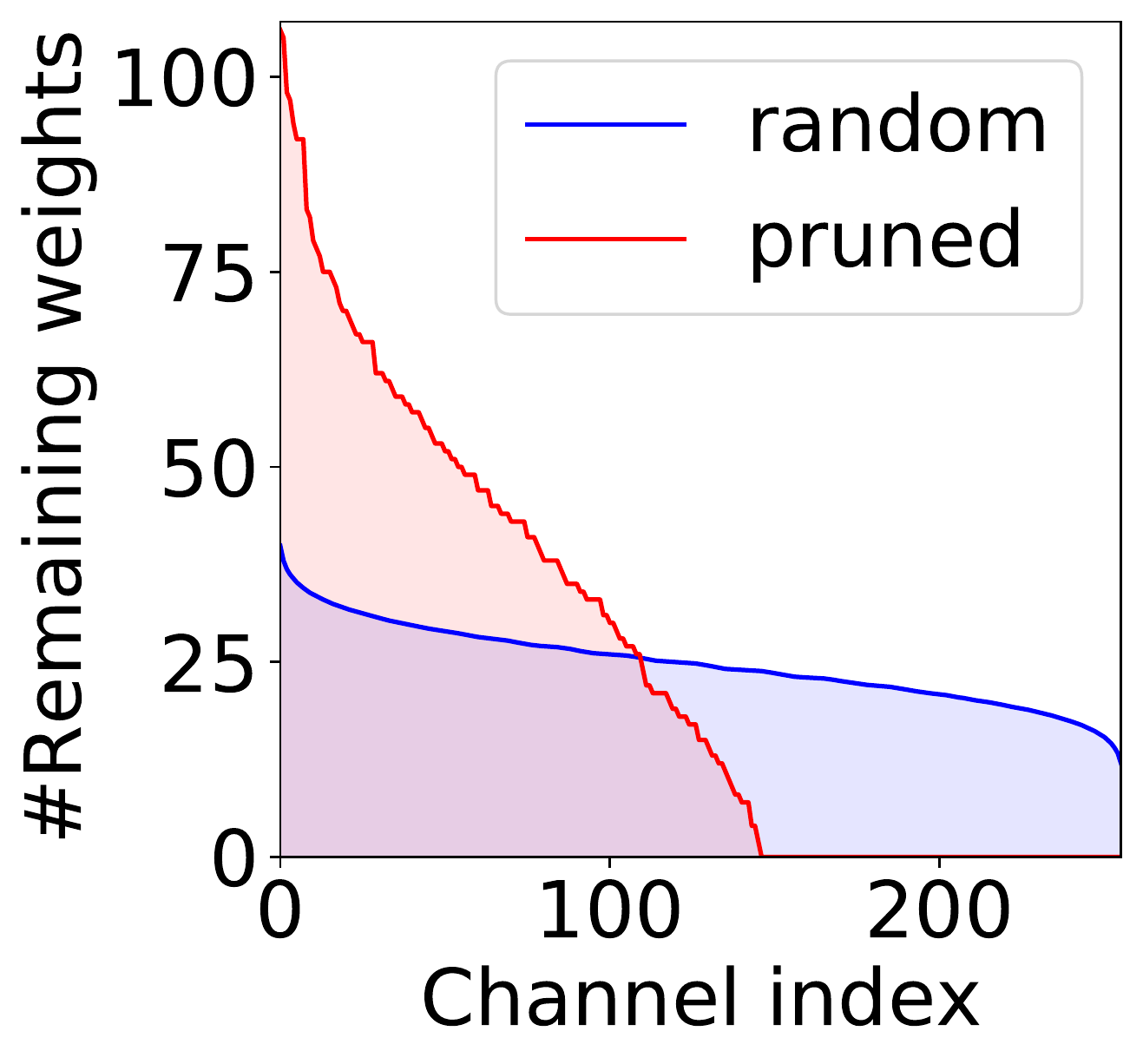}
    \caption{L1 Output Channel}
\end{subfigure}
~~
\begin{subfigure}{0.23\textwidth}
    \centering
    \includegraphics[width=\linewidth]{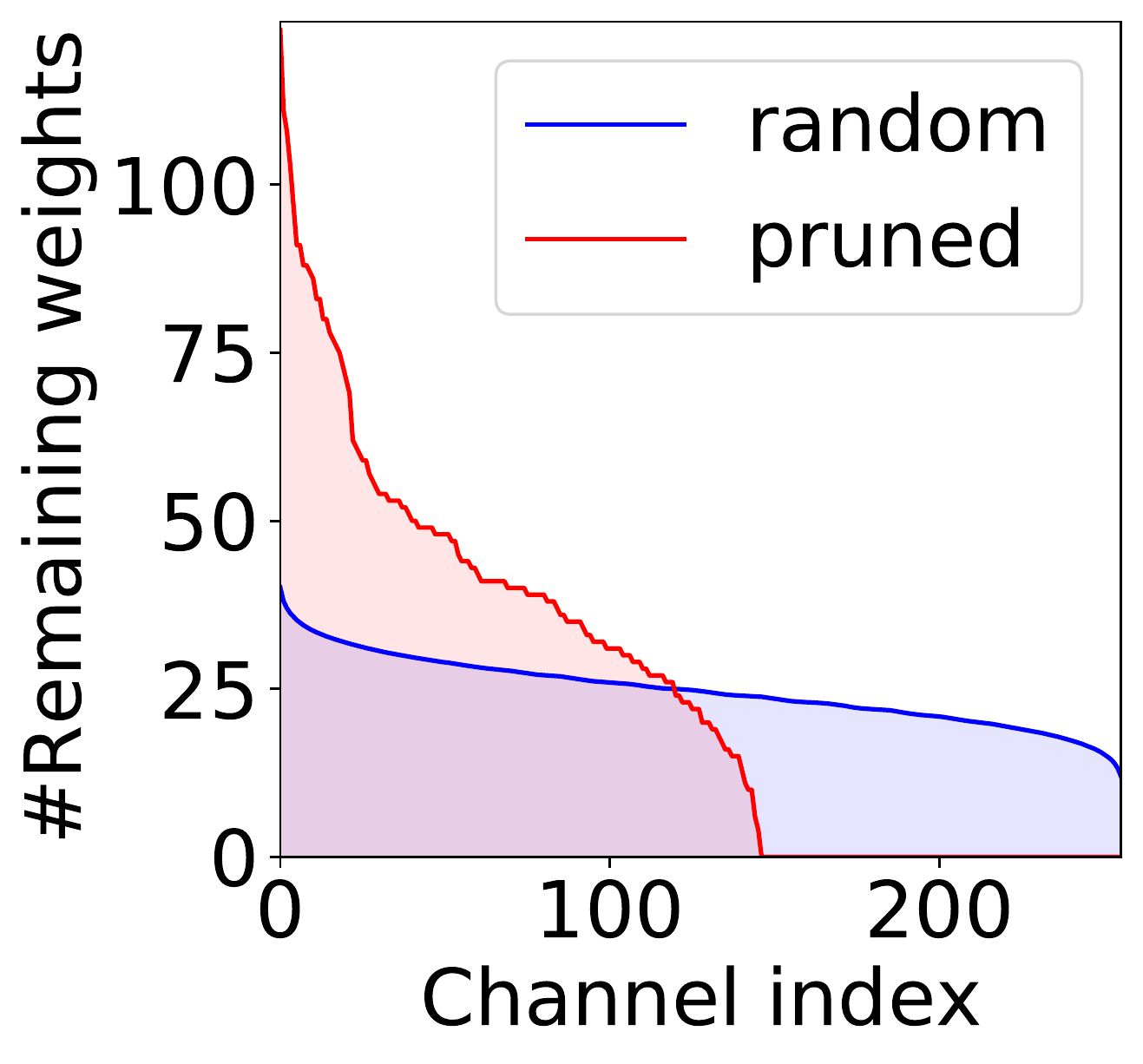}
    \caption{L2 Input Channel}
\end{subfigure}
~~
\begin{subfigure}{0.23\textwidth}
    \centering
    \includegraphics[width=\linewidth]{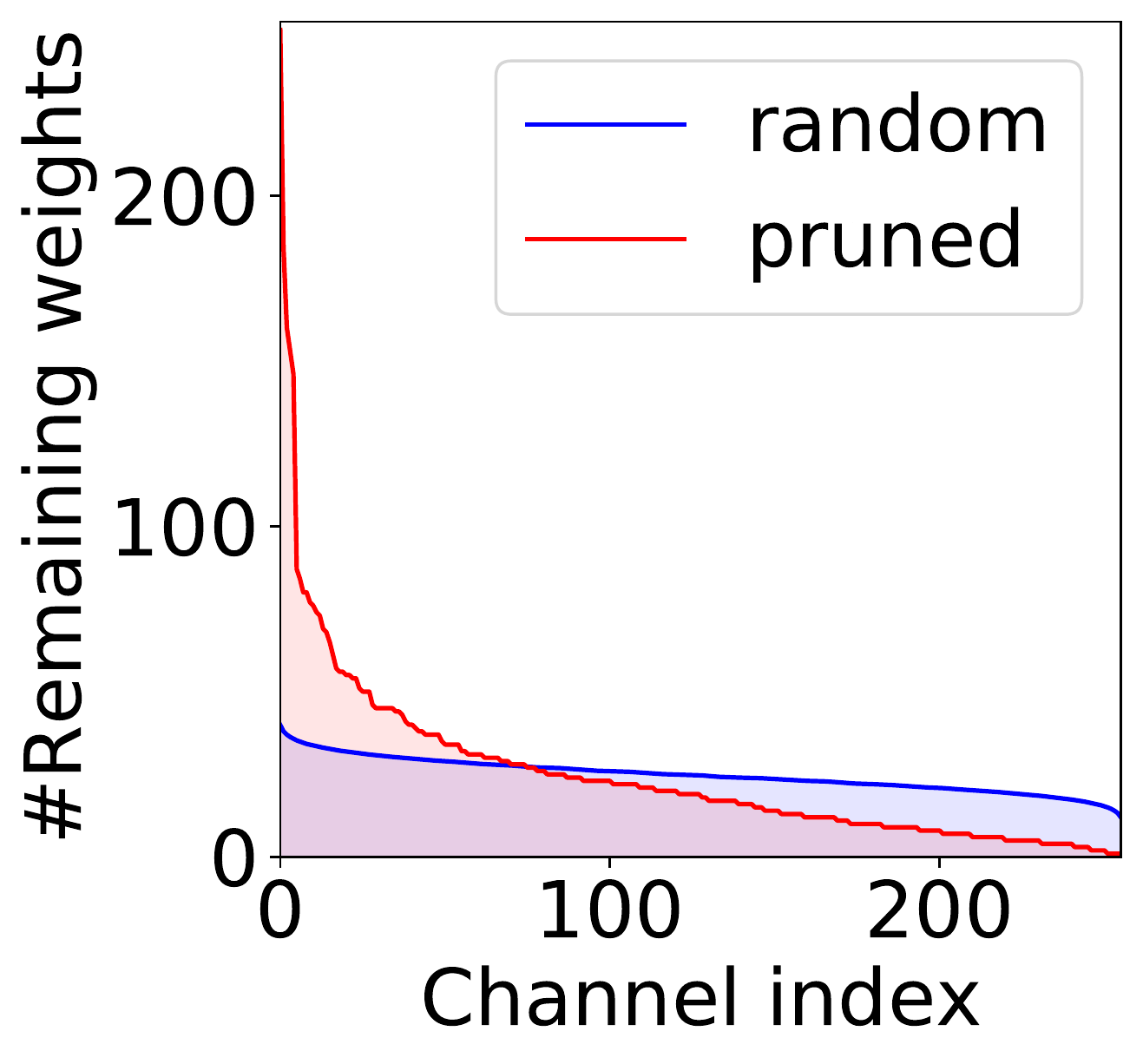}
    \caption{L2 Output Channel}
\end{subfigure}
\caption{Number of retained parameters in each input and output channel of layer1 \zzhao{(L1)} and layer2 \zzhao{(L2)} in the same residual block. We sort the numbers and plot the curves from the largest on the left to the smallest on the right. The red curves represent the mask obtained by our method; the blue curves depict what happens when randomly pruning the corresponding layer.}
\label{fig:mask_dist}
\end{figure}

Furthermore, we visualize two consecutive convolutional layers in the same residual block of the RN34 model.
We call them \textit{layer1} and \textit{layer2} following the forward pass.
In Figure~\ref{fig:mask_dist}, we plot the distribution of the retained parameters in each input channel and in each output channel, respectively.
We find that many output channels of layer1 and input channels of layer2, $40\%$ of all channels in this case, are totally pruned.
As a reference, we also plot the distribution of random pruning, based on the average of $500$ simulations.
\cliucamera{As demonstrated in Figure~\ref{fig:mask_dist}, the distribution of the retained parameters in each channel is much more uniform in this case.}
Our theoretical analysis in Appendix~\ref{subsec:random structure} demonstrates that, in a randomly pruned network, it is almost impossible to have even one entirely pruned channel.
\cliucamera{The comparison indicates that the mask $\rvm$ obtained by our method is structured.}

\begin{minipage}[b]{.65\textwidth}
\cliucamera{The observations in Figure~\ref{fig:mask_dist} also hold for kernels: a few kernels, of size $3 \times 3 $ \zzhaocamera{and thus having 9 entries}, are totally unpruned.
We show the distribution of the number of retained parameters in each kernel in Figure~\ref{fig:mask distribution 2} and provide the distribution by random uniform pruning as a reference.
Random uniform pruning yields no kernels with more than 3 retained parameters, but many such kernels can be observed in the masks generated by our method.}

Finally, in Figure~\ref{fig:mask 2 layers} of Appendix~\ref{subsec:mask}, \cliucamera{we visualize the positions of the pruned output channels of layer1 and the pruned input channels of layer2. We observe those pruned channels to be aligned.}
That is, some neurons representing both the output channels of layer1 and the input channels of layer2 are entirely removed.
We defer additional discussions, figures and results to Appendix~\ref{subsec:mask}.
The pattern of the structures learned by our method indicates the potential of regular pruning for a randomly-initialized network in the presence of adversarial attacks.
We leave this as future work.
\end{minipage}
\begin{minipage}[b]{0.05\textwidth}
\end{minipage}
\begin{minipage}[b]{.3\textwidth}
\includegraphics[width = \textwidth]{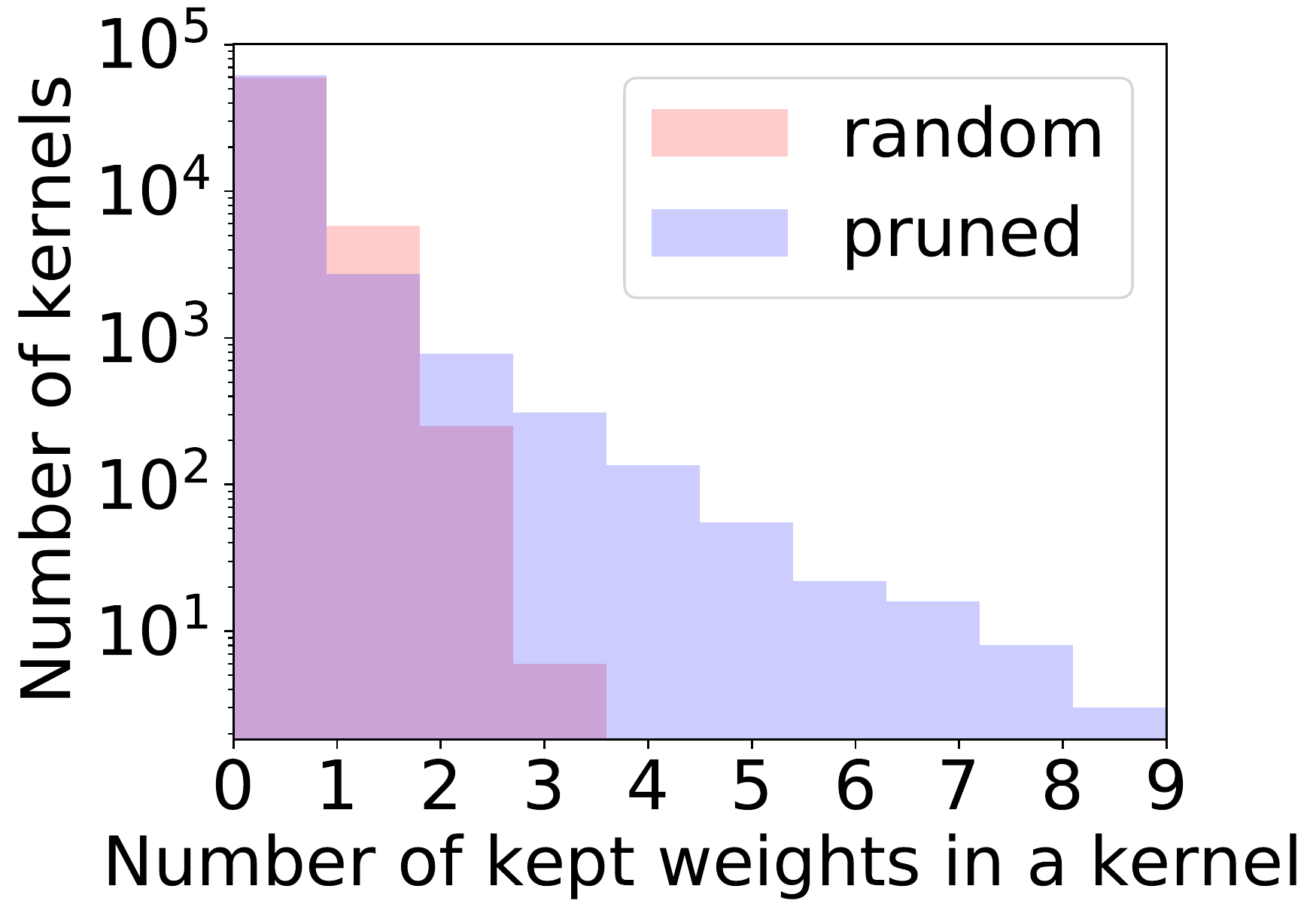}
\captionof{figure}{Distribution of the number of retained parameters in each kernel. The y-axis is in log-scale.} \label{fig:mask distribution 2}
\vspace{0.8cm}
\end{minipage}
\section{Conclusion}

We have proposed a method to obtain robust binary models by pruning randomly-initialized networks, thus extending the \textit{Strong Lottery Ticket Hypothesis} to the case of robust binary networks.
In contrast to the state-of-the-art methods, we learn the structure of robust subnetworks without updating the parameters.
Furthermore, we have proposed an \textit{adaptive pruning} strategy and last batch normalization layer to stabilize the training and improve performance.
Finally, we have relied on binary initialization to obtain more compact models.

Our extensive results on various benchmarks have demonstrated that our approach outperforms existing methods for training compressed robust models.
Furthermore, we have observed interesting structured patterns occurring in the parameters retained in the subnetworks.
This opens the door to further investigations on the structure of the robust subnetworks and on the design of regular pruning strategies in the adversarial scenario.


\section*{Checklist}


\begin{enumerate}

\item For all authors...
\begin{enumerate}
  \item Do the main claims made in the abstract and introduction accurately reflect the paper's contributions and scope?
    \answerYes{The paragraphs before the Notation part in the introduction demonstrate the findings and the contributions of this paper.}
  \item Did you describe the limitations of your work?
    \answerYes{This paper focuses on irregular pruning. We point out that we need hardware customization to fully take advantage of the efficiency improvement.}
  \item Did you discuss any potential negative societal impacts of your work?
    \answerNo{Our proposed algorithms are generic. This paper does not focus on a particular application.}
  \item Have you read the ethics review guidelines and ensured that your paper conforms to them?
    \answerYes{}
\end{enumerate}

\item If you are including theoretical results...
\begin{enumerate}
  \item Did you state the full set of assumptions of all theoretical results?
    \answerYes{All the theoretical results are put in the appendix.}
        \item Did you include complete proofs of all theoretical results?
    \answerYes{We provide detailed proofs for all theorems we claim.}
\end{enumerate}

\item If you ran experiments...
\begin{enumerate}
  \item Did you include the code, data, and instructions needed to reproduce the main experimental results (either in the supplemental material or as a URL)?
    \answerYes{It is in the supplemental material.}
  \item Did you specify all the training details (e.g., data splits, hyperparameters, how they were chosen)?
    \answerYes{We highlight key settings in the experiment part. More details are put in the appendix.}
        \item Did you report error bars (e.g., with respect to the random seed after running experiments multiple times)?
    \answerNo{We do not report the error bars in the main table that compares our method with baselines. As the ablation study in Table 2 shows, when we use the last batch normalization layer and adaptive pruning, the performance variance for different initializations is small, compared with the performance gap between our methods and baselines.}
        \item Did you include the total amount of compute and the type of resources used (e.g., type of GPUs, internal cluster, or cloud provider)?
    \answerYes{We put these details in the appendix.}
\end{enumerate}

\item If you are using existing assets (e.g., code, data, models) or curating/releasing new assets...
\begin{enumerate}
  \item If your work uses existing assets, did you cite the creators?
    \answerYes{}
  \item Did you mention the license of the assets?
    \answerYes{}
  \item Did you include any new assets either in the supplemental material or as a URL?
    \answerYes{}
  \item Did you discuss whether and how consent was obtained from people whose data you're using/curating?
    \answerNA{All the data we use is from popular benchmarks}
  \item Did you discuss whether the data you are using/curating contains personally identifiable information or offensive content?
    \answerNA{}
\end{enumerate}

\item If you used crowdsourcing or conducted research with human subjects...
\begin{enumerate}
  \item Did you include the full text of instructions given to participants and screenshots, if applicable?
    \answerNA{}
  \item Did you describe any potential participant risks, with links to Institutional Review Board (IRB) approvals, if applicable?
    \answerNA{}
  \item Did you include the estimated hourly wage paid to participants and the total amount spent on participant compensation?
    \answerNA{}
\end{enumerate}

\end{enumerate}


\newpage

\appendix

\section{Analysis}

\subsection{Analysis of Layerwise Pruning Strategies} \label{subsec:prune_strategy}

\subsubsection{Fixed Pruning Rate}

We consider an $L$-layer neural network and each layer has $n_1$, $n_2$, ..., $n_L$ parameters, we retrain $m_1$, $m_2$, ..., $m_L$ parameters after pruning.
As such, the total number of combinations $\Pi_{i = 1}^L \left(\begin{aligned}n_i \\ m_i \end{aligned}\right)$ is the size of search space of the subnetworks.
The following theorem shows, the \textit{fixed pruning rate} strategy is the strategy which approximates the maximization of the total number of combinations.

\begin{theorem} \label{thm:fix_pr}
Consider an $L$-layer neural network with $n_1$, $n_2$, ..., $n_L$ parameters in each layer, we retain $m_1$, $m_2$, ..., $m_L$ parameters after pruning.
Given a predefined pruning rate $r = 1 - \frac{\sum_{i = 0}^L m_i}{\sum_{i = 0}^L n_i}$, the optimal numbers of post-pruning parameters $\{m_i\}_{i = 1}^L$ that maximizing the total number of combinations $\Pi_{i = 1}^L \begin{pmatrix}n_i \\ m_i \end{pmatrix}$ satisfy the following inequality:
\begin{equation}
\begin{aligned} \label{eq:maxcombcond}
\forall 1 \leq j, k \leq L,\ \left|\frac{m_j}{n_j} - \frac{m_k}{n_k}\right| < \frac{1}{n_j} + \frac{1}{n_k}
\end{aligned}
\end{equation}
\end{theorem}

We defer the proof to Appendix~\ref{subsec:balanced}.
Specifically, we let $n_k$ in (\ref{eq:maxcombcond}) be the largest layer in the network without the loss of generality, we then have $\forall i \leq j \leq L, j \neq k$, $\left|m_j - \frac{m_k}{n_k}n_j\right| < \frac{n_j}{n_k} + 1 \leq 2$.
$m_j$ is the number of retained parameters and thus an integer, so Theorem~\ref{thm:fix_pr} indicates the pruning rate of each layer is close to each other when we aim to maximize the total number of combinations.
Therefore, we can consider the \textit{fixed pruning rate} strategy, i.e., $1 - r = \frac{m_1}{n_1} = \frac{m_2}{n_2} = ... = \frac{m_L}{n_L}$, as an approximation to maximize the total number of combinations.

\textbf{Drawbacks} While this strategy may seem intuitive, it does not take the differences in layer size into account.
In practice, the number of parameters in different layers can vary widely. For example, residual networks~\cite{he2016deep} have much fewer parameters in the first and last layers than in the middle ones.
Using \textit{fixed pruning rate} thus yields very few parameters after pruning within such small layers.
For example, when $r = 0.99$, only 17 parameters are left after pruning for a convolutional layer with $3$ input channels, $64$ output channels and a kernel size of $3$. Such a small number of parameters has two serious drawbacks: 1) It greatly limits the expression power of the network; 2) it makes the edge-popup algorithm less stable, because adding or removing a single parameter then has a large impact on the network's output. This instability becomes even more pronounced in the presence of adversarial samples, because the gradients of the model parameters are more scattered than when training on clean inputs~\cite{liu2020loss}.

\subsubsection{Fixed Number of Parameters}

To overcome drawbacks of the fixed pruning rate strategy, we study an alternative strategy aiming to maximize the total number of paths from the input to the output in the pruned network.
For a feedforward network, the total number of such paths is upper bounded by $\Pi_{i = 1}^L m_i$.
The following theorem demonstrates that the pruning strategy that maximizes this upper bound consists of retaining the same number of parameters in every layer, except for the layers that initially have too few parameters, for which all parameters should then be retained.
This optimal strategy is the \textit{fixed number of parameters} mentioned in Section~\ref{subsec:adaptivepruning}.

\begin{theorem} \label{thm:maxpath}
Consider an $L$-layer feedforward neural network with $n_1, n_2, ..., n_L$ parameters in its successive layers, from which we retain $m_1, m_2, ..., m_L$ parameters, respectively, after pruning. Given a predefined sparsity ratio $r = 1 - \frac{\sum_{i = 1}^L m_i}{\sum_{i = 1}^L n_i}$, the numbers of post-pruning parameters $\{m_i\}_{i = 1}^L$ that maximize the upper bound of the total number of the input-output paths $\Pi_{i = 1}^L m_i$ have the following property: $\forall 1 \leq j \leq L$, $m_j$ satisfies either of the following two conditions: 1) $m_j = n_j$; 2) $\forall  1 \leq k \leq L, m_j \geq m_k - 1$.
\end{theorem}

The two conditions in Theorem~\ref{thm:maxpath} mean we retain the same number of parameters for each layer except for ones totally unpruned.
We defer the proof to Appendix~\ref{subsec:maxpath}, where we use proof by contraction.

\textbf{Drawbacks} While this \textit{fixed number of parameters} strategy addresses the problem of obtaining too small layers arising in the \textit{fixed pruning rate} one, it suffers from overly emphasizing the influence of the small layers. That is, 
the smaller layers end up containing too many parameters.
In the extreme case, some layers are totally unpruned when the pruning rate $r$ is small.
This is problematic in our settings, since the model parameters are random and not updated.
The unpruned layers based on random parameters provide a large amount of noise in the forward process.
Furthermore, this strategy significantly sacrifices the expression power of the big layers. 




\subsection{Analysis of Acceleration by Binary Initialization} \label{subsec:binary_analysis}

In this section, we analyze the acceleration benefit of binary initialization.
Since most of the forward and backward computational complexity for the models studied in this paper is consumed by the ``Convolutional-BatchNorm-ReLU'' block, the acceleration rate on such blocks is a good approximation of that on the whole network.
Therefore, we concentrate on the ``Convolution-BatchNorm-ReLU'' block here.

For simplicity, we assume the feature maps and convolutional kernels are all squares.
Without the loss of generality, we consider $r_{in}$-channel input feature maps of size $s$, the size of the convolutional kernel is $c$ and the convolutional layer outputs $r_{out}$ channels.
Here, the binary layers represent the layer whose parameters are either $-1$ or $+1$.

\textbf{Forward Pass} For full-precision dense networks, the number of FLOP operations of the convolutional layer is $2c^2s^2r_{in}r_{out}$.
By contrast, the complexity can be reduced to $c^2s^2r_{in}r_{out}$ for binary dense layers, convolution operation with a binary kernel does not include any multiplication operations.
Correspondingly, for sparse layers whose pruning ratio is $r$, the complexity of full-precision sparse networks and of the binary sparse networks can be reduced to $2(1-r)c^2s^2r_{in}r_{out}$ and $(1-r)c^2s^2r_{in}r_{out}$, respectively.

The batch normalization layer will consume $3s^2r_{out}$ FLOP operations during inference and $10s^2r_{out}$ during training.
The additional operations during training are due to the update of running statistics.
Note that, the batch normalization layer in our random initialized network does not contain any trainable parameters, so there is no scaling parameters after normalization.
The ReLU layer will always consume $s^2r_{out}$ FLOP operations.

To sum up, we calculate the complexity ratio of the binary ``Convolution-BatchNorm-ReLU'' block over its full-precision counterpart in the forward pass.
For dense layers, the ratio is $\frac{c^2r_{in} + 11}{2c^2r_{in} + 11}$ for the training time and $\frac{c^2r_{in} + 4}{2c^2r_{in} + 4}$ for the inference.
For sparse layers, the ratio is $\frac{(1-r)c^2r_{in} + 11}{2(1-r)c^2r_{in} + 11}$ for the training time and $\frac{(1-r)c^2r_{in} + 4}{2(1-r)c^2r_{in} + 4}$ for the inference.

\textbf{Backward Pass} Compared with the forward pass, the backward pass has some computational overhead, because we need to calculate the gradient with respect to the score variable $\rvs$ associated with the convolutional kernels.
For both dense and sparse networks, the overhead is $2c^2s^2r_{in}r_{out} + c^2r_{in}r_{out}$ for full precision layers and $2c^2s^2r_{in}r_{out}$ for binary layers.
Note that, the overhead is independent of the pruning rate $r$ because the pruning function is treated as the identity function in the backward pass.
In addition, the difference here between the full precision layer and binary layer arises from the multiplication when we backprop the gradient through the weights.

To sum up, we calculate the complexity ratio of the binary ``Convolution-BatchNorm-ReLU'' block over its full-precision counterpart in the backward pass.
We only back propagate the gradient in the training time, so the batch normalization layer is always the training mode.
For dense layers, the ratio is $\frac{3c^2s^2r_{in} + 4s^2}{4c^2s^2r_{in} + 4s^2 + c^2r_{in}}$.
For sparse layers, the ratio is $\frac{3(1-r)c^2s^2r_{in} + 4s^2}{4(1-r)c^2s^2r_{in} + 4s^2 + c^2r_{in}}$.

\begin{table}[h]
\small
\centering
\setlength{\tabcolsep}{3pt}
\begin{tabular}{p{2.8cm}p{5.5cm}<{\centering}p{5.0cm}<{\centering}}
\Xhline{4\arrayrulewidth}
 & Full Precision & Binary \\
\hdashline
Forward - Training   & $2(1-r)c^2s^2r_{in}r_{out} + 11s^2r_{out}$ & $(1-r)c^2s^2r_{in}r_{out} + 11s^2r_{out}$ \\
Forward - Evaluation & $2(1-r)c^2s^2r_{in}r_{out} + 4s^2r_{out}$  & $(1-r)c^2s^2r_{in}r_{out} + 4s^2r_{out}$ \\
Backward - Training  & $4(1-r)c^2s^2r_{in}r_{out} + 4s^2r_{out} + c^2r_{in}r_{out}$ & $3(1-r)c^2s^2r_{in}r_{out} + 4s^2r_{out}$ \\
\Xhline{4\arrayrulewidth}
\end{tabular}
\caption{The complexity in FLOP operations of the sparse ``Convolution-BathNorm-ReLU'' block in both full precision and binary case. The pruning rate is $r$.}
\label{tbl:complexity}
\end{table}

\textbf{Discussion} We summarize the complexity in FLOP operations of the sparse ``Convolution-BatchNorm-ReLU'' block in different scenarios.
We can now conclude that compared with the full precision block, the binary block decrease the overall complexity in two places: 1) we save $(1-r)c^2s^2r_{in}r_{out}$ FLOPs for the convolution and transpose convolution operations in the forward and backward pass, respectively; 2) for the backpropagation, we save $c^2r_{in}r_{out}$ FLOPs, because there is no multiplication when we backprop the gradient through the weights for binary blocks.

We consider the practical settings: $r = 0.99$, $c = 3$, $r_{in} = r_{out} = 128$, $s = 16$.
The complexity ratio of the binary block over the full precision block in the forward pass is $\frac{(1-r)c^2r_{in} + 11}{2(1-r)c^2r_{in} + 11} = 0.6616$ for the training mode and $\frac{(1-r)c^2r_{in} + 4}{2(1-r)c^2r_{in} + 4} = 0.5740$ for the evaluating mode, respectively.
The complexity ratio in the backward pass is $\frac{3(1-r)c^2s^2r_{in} + 4s^2}{4(1-r)c^2s^2r_{in} + 4s^2 + c^2r_{in}} = 0.7065$.
That is to say, compared with the full precision block, the binary block under this setting can save around $34\%$ and $29\%$ time in the forward and backward passes during training; for inference, it can save $43\%$ time.


\subsection{Analysis of the Normalization Layer before Softmax} \label{subsec:normalization}

We consider a $L$-layer neural network and each layer has $l_1$, $l_2$, ..., $l_L$ neurons.
Let $\rvu \in \sR^{l_{L - 1}}$, $\rmW \in \sR^{l_L \times l_{L -1}}$, $\rvo \in \sR^{l_L}$ be the output of the penultimate's output, the weight matrix of the last fully-connected layer and the last layer's output, respectively.
In addition, we use $c \in \{1, 2, ..., l_L\}$ to denote the label of the data and omit the bias term of the last layer since it is initialized as $0$ and is not updated.
For the 1-dimensional batch normalization layer, we use $\rvb \in \sR^{l_L}$ and $\rvv \in \sR^{l_L}$ to represent the running mean and running standard deviation, respectively.

Therefore, the loss objective $\gL_{wo}$ and its gradient of the model without the 1-dimensional batch normalization layer is:
\begin{equation}
\begin{aligned}
\gL_{wo} &= - log \frac{e^{\rvo_c}}{\sum_{i = 1}^{l_L} e^{\rvo_i}} \\
\frac{\partial \gL_{wo}}{\partial \rvo_j} &= \frac{e^{\rvo_j}}{\sum_{i = 1}^{l_L} e^{\rvo_i}} - \mathbf{1}(j = c)
\end{aligned}
\end{equation}

Correspondingly, the loss objective $\gL_{wi}$ and its gradient of the model with the 1-dimensional batch normalization layer is:
\begin{equation}
\begin{aligned}
\gL_{wi} &= - log \frac{e^{(\rvo_c - \rvb_c) / \rvv_c}}{\sum_{i = 1}^{l_L} e^{(\rvo_i - \rvb_i) / \rvv_i}} \\
\frac{\partial \gL_{wi}}{\partial \rvo_j} &= \frac{1}{\rvv_j} \left( \frac{e^{(\rvo_c - \rvb_c) / \rvv_c}}{\sum_{i = 1}^{l_L} e^{(\rvo_i - \rvb_i) / \rvv_i}} - \mathbf{1}(j = c) \right)
\end{aligned}
\end{equation}

Now we consider the case when the model parameter $\rmW$ is multiplied by a factor $\alpha > 1$: $\rmW' = \alpha \rmW$ and assume the output of the penultimate layer is unchanged.
In practice, $\alpha$ is far more than $1$.
For example, if the penultimate layer has $512$ neurons, $\alpha$ will be $16$ when we change kaiming constant initialization to binary initialization.
Based on this, the new output of the last layer is $\rvo' = \alpha \rvo$.
For the model with the normalization layer, the new statistics are $\rvb' = \alpha \rvb$ and $\rvv' = \alpha \rvv$.
In this regard, we can then recalculate the gradient of the loss objective as follows:

\begin{equation}
\begin{aligned}
\frac{\partial \gL'_{wo}}{\partial \rvo'_j} &= \frac{e^{\rvo'_j}}{\sum_{i = 1}^{l_L} e^{\rvo'_i}} - \mathbf{1}(j = c) = \frac{e^{\alpha \rvo_j}}{\sum_{i = 1}^{l_L} e^{\alpha \rvo_i}} - \mathbf{1}(j = c) \\
\frac{\partial \gL'_{wi}}{\partial \rvo'_j} &= \frac{1}{\rvv'_j} \left( \frac{e^{(\rvo'_c - \rvb'_c) / \rvv'_c}}{\sum_{i = 1}^{l_L} e^{(\rvo'_i - \rvb'_i) / \rvv'_i}} - \mathbf{1}(j = c) \right) = \frac{1}{\alpha \rvv_j} \left( \frac{e^{(\rvo_c - \rvb_c) / \rvv_c}}{\sum_{i = 1}^{l_L} e^{(\rvo_i - \rvb_i) / \rvv_i}} - \mathbf{1}(j = c) \right)
\end{aligned} \label{eq:alphaloss}
\end{equation}

We first study the case without the normalization layer.
The first term $\frac{e^{\alpha \rvo_j}}{\sum_{i = 1}^{l_L} e^{\alpha \rvo_i}}$ of the gradient $\frac{\partial \gL'_{wo}}{\partial \rvo'_j}$ converge to $\mathbf{1}(j = \mathrm{argmax}_i \rvo_i)$ exponentially.
For correctly classified inputs, $\frac{\partial \gL'_{wo}}{\partial \rvo'_j}$ converge to $0$ exponentially with $\alpha$.
In addition, the gradient $\frac{\partial \gL'_{wo}}{\partial \rvu} = \rmW'^T \frac{\partial \gL'_{wo}}{\partial \rvo'_j} = \alpha \rmW^T \frac{\partial \gL'_{wo}}{\partial \rvo'_j}$ also vanish with $\alpha$.
$\frac{\partial \gL'_{wo}}{\partial \rvu}$ is backward to previous layers, leading to gradient vanishing.
For incorrectly classified inputs, $\frac{\partial \gL'_{wo}}{\partial \rvo'_j}$ converge to $\mathbf{1}(j = \mathrm{argmax}_i \rvo_i) - \mathbf{1}(j = c)$, which is a vector with $c$-th element being $-1$, the element corresponding to the output label being $+1$ and the rest elements being $0$.
In this case, the gradient backward $\frac{\partial \gL'_{wo}}{\partial \rvu} = \alpha \rmW^T \frac{\partial \gL'_{wo}}{\partial \rvo'_j}$ will be approximately multiplied by $\alpha$, causing gradient exploding.

By contrast, in the case of the model with the normalization layer, $\frac{\partial \gL'_{wi}}{\partial \rvo'_j} = \frac{1}{\alpha} \frac{\partial \gL_{wi}}{\partial \rvo_j}$.
The factor $\frac{1}{\alpha}$ is cancelled out when we calculate $\frac{\partial \gL'_{wi}}{\partial \rvu} = \rmW'^T \frac{\partial \gL'_{wi}}{\partial \rvo} = \rmW^T \frac{\partial \gL_{wi}}{\partial \rvo}$.
This means the gradient backward remains unchanged if we use the 1-dimensional batch normalization layer, which maintains the stability of training if we scale the model parameters.

To conclude, the 1-dimensional batch normalization layer is crucial to maintain the stability of training if we use binary initialization.
Without this layer, the training will suffer from gradient vanishing for correctly classified inputs and gradient exploding for incorrectly classified inputs.

\subsection{Analysis of the structure of a randomly pruned network}
\label{subsec:random structure}

In this section, we provide preliminary analysis of the structure of a randomly pruned network.

As a starting point, we first estimate the probability of $k$ retained parameters in a $3 \times 3$ kernel.
Given the pruning rate $r_i$ for the layer $i$ with $n_i$ weights, the number of the retained parameters is $m_i \defeq (1-r_i)n_i$. 
We assume $m_i$ lies in a proper range: $9 \ll m_i < \frac{1}{9}n_i$.
This is true when $n_i$ is large and $r_i > \frac{8}{9}$.

For each kernel $j$, we use $X_j$ to represent its number of retained parameters.
It is difficult to calculate $P(X_j = k)$ directly because $\{X_j\}_j$ are constrained by: 1) $\sum_{j} X_j = m_i$; 2) $\forall j, 0 \leq X_j \leq 9 $.
However, in the case of random pruning, we have $E[X_j] = \frac{9m_i}{n_i} = 9(1-r_i) < 1$.
In this regard, we can make the approximation by removing the constraint $X_j \leq 9$.

Therefore, we can reformulate the problem of calculating $P(X_j = k)$ as:
\textit{Given $m_i$ steps, randomly select one box out of the total $\frac{n_i}{9}$ boxes and put one apple in it. $P(X_j = k)$ is then the probability for the box $j$ to have $k$ apples.}

In this approximation, it is straightforward to have $P(X_j=k) = \binom{m_i}{k}P_i^k \cdot (1-P_i)^{m_i-k}, 0\leq k \leq 9$ where $P_i = \frac{9}{n_i}$.
Based on the assumption that $n_i$ is large, $P_i \approx 0$.
Therefore, $P(X_j = 0) = (1-\frac{9}{n_i})^{m_i} \approx e^{9(1-r_i)}$.
For $k > 1$, we apply Stirling approximation $n! \approx \sqrt{2\pi n}(\frac{n}{e})^n$ to the binomial coefficient, then

\begin{equation}
    \begin{aligned}
    P(X_j=k) &\approx \frac{m_i^{m_i+0.5}}{\sqrt{2\pi} k^{k+0.5} (m_i-k)^{m_i-k+0.5}} \cdot (\frac{9}{n_i})^k \cdot (1-\frac{9}{n_i})^{m_i-k} \\
    &=\sqrt{\frac{m_i}{2\pi k (m_i-k)}} \cdot (\frac{9(1-r_i)}{k})^k \cdot (1 + \frac{k}{m_i-k})^{m_i-k} \cdot (1-\frac{9}{n_i})^{m_i-k}\\
    \end{aligned}
    \label{eq:kernel distribution}
\end{equation}

The second equality is based on the fact $m_i = (1 - r_i)n_i$.
Since $m_i \gg 9 > k$ by the assumption and $n_i \gg m_i$, we can approximate $(1-\frac{9}{n_i})^{m_i-k}$ to $1-\frac{9(m_i-k)}{n_i} \approx 1$, then

\begin{equation}
    \begin{aligned}
    P(X_j=k) \approx \sqrt{\frac{m_i}{2\pi k (m_i-k)}} \cdot (\frac{9e(1-r_i)}{k})^k
    =\sqrt{\frac{m_i}{2\pi k (m_i-k)}} \cdot (\frac{c}{k})^k\\
    \end{aligned}
\end{equation}

where $c=9e(1-r_i)$ is a constant.

As shown in the equation above, $P(X_j = k)$ decreases drastically when $k$ increases. Therefore, in a randomly pruned layer $i$ with $n_i=3\times3\times256\times256=589824$ and $r_i=0.99$, it is almost impossible to see kernels who have at least 4 retained parameters, because according to the above formula, the estimated number of kernels in that layer having 3 retained parameters is $\frac{n_i}{9}\times P(X_j=3) \approx 8.19$, and the number of kernels having 4 retained parameters is $\approx$ 0.18.

Now we consider the number of retained parameters in a channel.
For the layer of $r_{in}$ input channels and $r_{out}$ output channels, it has $r_{in}\times r_{out}\times 3\times 3$ parameters.
We use $Y_j$ to represent the number of the retained parameters for the input channel $j$.
Similarly, for the random pruning, we have

\begin{equation}
    \begin{aligned}
    P(Y_j = k)\approx \sqrt{\frac{m_i}{2\pi k (m_i-k)}} \cdot (\frac{c'}{k})^k \cdot (1-\frac{1}{r_{in}})^{m_i-k}
    \end{aligned}
\end{equation}

where $c' = 9er_{out}(1-r_i)$ is a constant.

By plotting the distribution $P(Y_j)$, it is easy to find that the distribution of $Y_j$ concentrates around the neighborhood of $k = \frac{m_i}{a}$, and decreases significantly as $Y_j$ deviates from it.
\section{Proofs of Theoretical Results}

\subsection{Proof of Theorem~\ref{thm:fix_pr}} \label{subsec:balanced}

\begin{proof}
We pick arbitrary $0 < j, k \leq L$ and generates two sequences $\{\widehat{m}_i\}_{i = 1}^L$, $\{\widetilde{m}_i\}_{i = 1}^L$ as follows:
\begin{equation}
\begin{aligned}
\widehat{m}_j = m_j - 1, \widehat{m}_k = m_k + 1, \widehat{m}_i = m_i \forall i \neq j, i \neq k. \\
\widetilde{m}_j = m_j + 1, \widetilde{m}_k = m_k - 1, \widetilde{m}_i = m_i \forall i \neq j, i \neq k. 
\end{aligned}
\end{equation}

Consider $\{m_i\}_{i = 1}^L$ the optimality that maximizes the combination number $\Pi_{i = 1}^L \begin{pmatrix} n_i \\ m_i \end{pmatrix}$.
We have the following inequality:
\begin{equation}
\begin{aligned}
1 > \frac{\Pi_{i = 1}^L \begin{pmatrix} n_i \\ \widehat{m}_i \end{pmatrix}}{\Pi_{i = 1}^L \begin{pmatrix} n_i \\ m_i \end{pmatrix}} = \frac{m_j}{n_j - m_j + 1} \frac{n_k - m_k}{m_k + 1} \\
1 > \frac{\Pi_{i = 1}^L \begin{pmatrix} n_i \\ \widetilde{m}_i \end{pmatrix}}{\Pi_{i = 1}^L \begin{pmatrix} n_i \\ m_i \end{pmatrix}} = \frac{n_j - m_j}{m_j + 1} \frac{m_k}{n_k - m_k + 1}
\end{aligned}
\end{equation}

Reorganize the inequalities above, we obtain:

\begin{equation}
\begin{aligned}
- \left( \frac{1}{n_k} + \frac{m_k - m_j + 1}{n_j n_k}\right) < \frac{m_k}{n_k} - \frac{m_j}{n_j} < \left(\frac{1}{n_j} + \frac{m_j - m_k + 1}{n_j n_k}\right)
\end{aligned}
\end{equation}

Consider $1 \leq m_j \leq n_j$ and $1 \leq m_k \leq n_k$, we have $\frac{m_k - m_j + 1}{n_j n_k} \leq \frac{1}{n_j}$ and $\frac{m_j - m_k + 1}{n_j n_k} \leq \frac{1}{n_k}$.
As a result, we have the following inequality:

\begin{equation}
\begin{aligned}
\forall j, k, - \left( \frac{1}{n_j} + \frac{1}{n_k}\right) < \frac{m_k}{n_k} - \frac{m_j}{n_j} < \left(\frac{1}{n_j} + \frac{1}{n_k}\right)
\end{aligned}
\end{equation}

This concludes the proof.
\end{proof}

\subsection{Proof of Theorem~\ref{thm:maxpath}} \label{subsec:maxpath}

\begin{proof}
We proof the theorem by contradictory.
We assume the optimal $\{m_i\}_{i = 1}^L$ does not satisfy the property mentioned in Theorem~\ref{thm:maxpath}.
This means $\exists 1 \leq j \leq L$ such that $m_j < n_j$ and $\exists 1 \leq k \leq L, m_j < m_k - 1$.
Based on this, we then construct a new sequence $\{\widehat{m}_i\}_{i = 1}^L$ as follows:
\begin{equation}
\begin{aligned}
\widehat{m}_j = m_j + 1; \widehat{m}_k = m_k - 1; \forall i \neq j, i \neq k, \widehat{m}_i = m_i.
\end{aligned}
\end{equation}

We then calculate the ratio of $\Pi_{i = 1}^L \widehat{m}_i$ and $\Pi_{i = 1}^L m_i$:

\begin{equation}
\begin{aligned} \label{eq:con}
\frac{\Pi_{i = 1}^L \widehat{m}_i}{\Pi_{i = 1}^L m_i} = \frac{(m_j + 1)(m_k - 1)}{m_j m_k} = 1 + \frac{m_k - m_j - 1}{m_j m_k} > 1
\end{aligned}
\end{equation}

The last inequality is based on the assumption $m_j < m_k - 1$.
(\ref{eq:con}) indicates $\Pi_{i = 1}^L \widehat{m}_i > \Pi_{i = 1}^L m_i$, which contradicts the optimality of $\{m_i\}_{i = 1}^L$.
\end{proof}
\section{Algorithm} \label{sec:app_algorithm}

We provide the pseudo-code of the edge pop-up algorithm for adversarial robustness as Algorithm~\ref{alg:edgepopup}.
We use PGD to generate adversarial attacks.
$\Pi_{\gS_\epsilon}$ mean projection into the set $\gS_\epsilon$.

\begin{algorithm}[h]
\begin{algorithmic}
\STATE \textbf{Input:} training set $\gD$, batch size $B$, PGD step size $\alpha$ and iteration number $n$, adversarial budget $\gS_\epsilon$, pruning rate $r$, mask function $M$, the optimizer.
\STATE Random initialize the model parameters $\rvw$ and the scores $\rvs$.
\FOR {Sample a mini-batch $\{\rvx_i, y_i\}_{i = 1}^B \sim \gD$}
    \FOR {i = 1, 2, ..., B}
        \STATE Sample a random noise $\delta$ within the adversarial budget $\gS_\epsilon$.
        \STATE $\rvx_i^{(0)} = \rvx_i + \delta$
        \FOR {j = 1, 2, ..., n}
            \STATE $\rvx_i^{(j)} = \rvx_i^{(j - 1)} + \alpha \triangledown_{\rvx_i^{(j - 1)}} \gL(f(\rvw \odot M(\rvs, r), \rvx_i^{(j - 1)}), y_i)$
            \STATE $\rvx_i^{(j)} =  \rvx_i + \Pi_{\gS_\epsilon} \left( \rvx_i^{(j)} - \rvx_i \right)$
        \ENDFOR
    \ENDFOR
    \STATE Calculate the gradient $\rvg = \frac{1}{B} \sum_{i = 1}^B \triangledown_{\rvs} \gL(f(\rvw \odot M(\rvs, r), \rvx_i^{(n)}), y_i)$
    \STATE Update the score $\rvs$ using the optimizer.
\ENDFOR
\STATE \textbf{Output:} the pruning mask $M(\rvs, r)$.
\end{algorithmic}
\caption{Edge pop-up algorithm for adversarial robustness.} \label{alg:edgepopup}
\end{algorithm}

\change{We provide the pseudo-code of our algorithm on the ImageNet100 \zzhao{as Algorithm~\ref{alg:imagenet100}}. It incorporates FGSM~\cite{wong2020fast} with ATTA~\cite{chen2021efficient}.
In addition, due to the high resolution and large size of the ImageNet100 dataset, we need to compress the initial perturbation directory to reduce the overhead of memory consumption.
Here, we choose to downsample the original perturbation to reduce its resolution for storage, and then upsample it back to the original resolution when using it as the initial perturbation.
}

\begin{algorithm}[h]
\begin{algorithmic}
\STATE \textbf{Input:} training set $\gD$, batch size $B$, FGSM step size $\alpha$, adversarial budget $\gS_\epsilon$, pruning rate $r$, mask function $M$, the optimizer.
\STATE Random initialize the model parameters $\rvw$ and the scores $\rvs$.
\STATE Initialize the instance-to-perturbation dictionary $\gM = \{\}$
\FOR {Sample a mini-batch $\{\rvx_i, y_i\}_{i = 1}^B \sim \gD$}
    \FOR{$i = 1, 2, ..., n$}
        \STATE Data augmentation $\rvx_i \leftarrow A(\rvx_i)$
        \IF{$\rvx_i$ in $\gM$}
            \STATE Get the downsampled perturbation: $\delta'_i = A(\gM(\rvx_i))$
            \STATE Upsample $\delta'$ to the original resolution and get $\delta_i$.
        \ELSE
            \STATE Sample a random noise $\delta_i$ within the adversarial budget $\gS_\epsilon$
        \ENDIF
        \STATE $\delta_i \leftarrow \delta_i + \alpha \triangledown_{\delta_i} \gL(f(\rvw \odot M(\rvs, r), \rvx_i + \delta_i), y_i)$
        \STATE $\delta_i \leftarrow \Pi_{\gS_\epsilon} \delta_i$
        \STATE Update the dictionary by the downsampled perturbation $\delta'_i$: $\gM(\rvx_i) = A^{-1} (\delta'_i)$
    \ENDFOR
\ENDFOR
\STATE Calculate the gradient $\rvg = \frac{1}{B} \sum_{i = 1}^B \triangledown_{\rvs} \gL(f(\rvw \odot M(\rvs, r), \rvx_i + \delta_i), y_i)$
\STATE Update the score $\rvs$ using the optimizer.
\STATE \textbf{Output:} the pruning mask $M(\rvs, r)$.
\end{algorithmic}
\caption{\change{Accelerated training for ImageNet100.}} \label{alg:imagenet100}
\end{algorithm}
\section{Experiments}

\subsection{Experimental Settings} \label{subsec:exp_setting}

\textbf{General} The RN34 architecture we use in this paper is the same as the one in~\cite{ramanujan2020s, wang2020hydra}, and it has 21265088 trainable parameters.
The bias terms of all linear layers are initialized 0, and are thus disabled.
We also disable the learnable affine parameters in batch normalization layers, following the setup of~\cite{ramanujan2020s}.
\zzhao{Unless specified, the number of training epochs for CIFAR10 and CIFAR100 is 400, and for ImageNet100 there are 100 training epochs.}
The adversarial budget in this paper is based on $l_\infty$ norm and the perturbation strength $\epsilon$ is $8 / 255$ for CIFAR10,  $4/255$ for CIFAR100 and $2/255$ for ImageNet100.
\change{The resolution of CIFAR10 and CIFAR100 is $32 \times 32$; the resolution of ImageNet100 is $224 \times 224$.
ImageNet100 is a subset of ImageNet which consists of $100$ classes.}
\zzhao{The selection of these classes follows the settings of a python library called Continuum~\cite{douillardlesort2021continuum}.}
The PGD attacks used in our experiments have $10$ iterations and the step size is one-quarter of the $\epsilon$, respectively.
The AutoAttack (AA) consists of the following four attacks: 1) the untargeted $100$-iteration AutoPGD based on cross-entropy loss; 2) the targeted $100$-iteration AutoPGD based on difference of logits ratio (DLR) loss; 3) the targeted $100$-iteration FAB attack~\cite{croce2020minimally}; 4) the black-box $5000$-query Square attack~\cite{andriushchenko2020square}.
We use the same hyper parameters in all these component attacks as in the original AutoAttack implementation.\footnote{AutoAttack: \url{https://github.com/fra31/auto-attack}.}

We train the model using an SGD optimizer, with the momentum factor being $0.9$ and the weight decay factor being $5 \times 10^{-4}$.
The learning rate is initially $0.1$ and decays following the cosine annealing scheduler.
Finally, since adversarial training suffers from severe overfitting~\cite{rice2020overfitting}, we use a validation set consisting of $2\%$ of the training data to select the best model during training.

\textbf{Adversarial Training} We apply the same settings as above to adversarial training, except the choice of optimizer and learning rate. For full precision networks, we use an SGD optimizer with an initial learning rate of 0.1 and decreases by a factor of 10 in the 200th and 300th epoch for CIFAR10 and CIFAR100 models. 
\zzhao{For ImageNet100, the learning rate decreases by a factor of 10 in the 50th and 75th epoch.} For binary networks, we use Adam optimizer~\cite{kingma2014adam} suggested in~\cite{courbariaux2015binaryconnect} and have a cosine annealing learning rate schedule with an initial learning rate of $1\times 10^{-4}$.

\textbf{Baselines (FlyingBird(+), BCS, RST, HYDRA, ATMC)} Our results on the baselines are based on their original public implementation except that we use the validation set to pick the best model during training.
\zzhao{FlyingBird(+), BCS, and HYDRA do not inherently support binary networks, so we plug in the \textit{BinaryConnect} algorithm~\cite{courbariaux2015binaryconnect} with the same settings as the ones in adversarial training. We also plug in Algorithm~\ref{alg:imagenet100} for fast training on ImageNet100. We scale down the number of training epochs of FlyingBird(+), BCS, RST to 100 epochs, and HYDRA to 110 epochs (50 pretrain + 10 prune + 50 finetune). For ATMC, we use 50 epochs for each of the four training phases, adding up to 200 epochs in total.} 
In \zzhao{all baselines except RST}, the batch normalization layers in the model have affine operations and are learnable.
This introduces additional trainable parameters and is different from the network used in our method.

\textbf{Smaller RN34 Variants}
Based on the adaptive pruning strategy, we designed several smaller RN34 variants with approximately the same number of parameters as the pruned networks. 
These variants have the same topology as RN34 but have fewer channels in each layer.
In Table~\ref{tbl:smallnet}, we provide architecture details based on different values of $p$ when the pruning rate $r$ is $0.99$.
The \textit{Small RN34} model in Table~\ref{tbl:main comparison} represents the small model with $p = 0.1$ (\textit{Small RN34-p0.1} in Table~\ref{tbl:smallnet}), since it has better performance than the other small networks.

\begin{table}[ht]
\centering
\begin{tabular}{p{3cm}<{\centering}p{3.5cm}<{\centering}p{3.5cm}<{\centering}}
    \Xhline{4\arrayrulewidth}
    layer name & Small RN34-p0.1 & Small RN34-p1.0 \\
    \hline
    conv1 & 
    $3 \times 3$, 23 &
    $3 \times 3$, 6 \\
    \rule{0pt}{2em}
    Block1 &
    \bsplitcell{3 $\times$ 3, 23  \\ 3 $\times$ 3, 23 }$\times$ 3 & 
    \bsplitcell{3 $\times$ 3, 6  \\ 3 $\times$ 3, 6 }$\times$ 3 \\[1em]
    \rule{0pt}{2em}
    Block2 &
    \bsplitcell{3 $\times$ 3, 25  \\ 3 $\times$ 3, 25 }$\times$ 4 & 
    \bsplitcell{3 $\times$ 3, 13  \\ 3 $\times$ 3, 13 }$\times$ 4 \\[1em]
    \rule{0pt}{2em}
    Block3 &
    \bsplitcell{3 $\times$ 3, 27  \\ 3 $\times$ 3, 27 }$\times$ 6 & 
    \bsplitcell{3 $\times$ 3, 26  \\ 3 $\times$ 3, 26 }$\times$ 6 \\[1em]
    \rule{0pt}{2em}
    Block4 &
    \bsplitcell{3 $\times$ 3, 29  \\ 3 $\times$ 3, 29 }$\times$ 3 & 
    \bsplitcell{3 $\times$ 3, 51  \\ 3 $\times$ 3, 51 }$\times$ 3 \\[1em]
    \multicolumn{3}{c}{average pool, 10d-fc, softmax}\\ 
    \hline
    \#params & 201078 & 216360 \\
    \hline
\end{tabular}
\caption{RN34 variants that have similar layer sizes as the pruned RN34 obtained by different $p$ values. $3\times 3 \times 23$ means the kernel size is $3\times 3$ and there are 23 output channels.}
\label{tbl:smallnet}
\end{table}

\subsection{Additional Experimental Results}
\label{subsec: more results}

\subsubsection{Ablation Study in the Non-Adversarial Case} \label{subsec:vanilla}

In the non-adversarial case, we train the models using clean inputs and report the clean accuracy in Table~\ref{tbl:vanilla_ablation}.
Other hyper-parameters here are the same as in Table~\ref{tbl:main_ablation}.
Our conclusions from Table~\ref{tbl:main_ablation} also hold true here: the \textit{binary initialization} can achieves comparable performance as the \textit{Signed Kaiming Constant}; the last batch normalization layer helps improve performance for both initialization schemes.

\begin{table}[ht]
\centering
\begin{tabular}{p{2.1cm}<{\centering}p{2.4cm}<{\centering}p{2.4cm}<{\centering}p{2.4cm}<{\centering}p{2.4cm}<{\centering}}
\Xhline{4\arrayrulewidth}
\multirow{2}{*}{Prune Scheme} & \multicolumn{2}{c}{Signed KC} & \multicolumn{2}{c}{Binary} \\
& no LBN & LBN & no LBN & LBN \\
\hline
$p = 0.0$ & 93.25 & 93.99 & 93.64 & \textbf{94.05} \\
$p = 0.1$ & 92.12 & 93.98 & \textbf{93.84} & 93.99 \\
$p = 0.2$ & 92.96 & \textbf{94.35} & 89.27 & 93.87 \\
$p = 0.5$ & \textbf{93.44} & 94.29 & 90.85 & 94.00 \\
$p = 0.8$ & 90.93 & 92.57 & 90.37 & 92.42 \\
$p = 0.9$ & 91.31 & 92.26 & 90.51 & 90.12 \\
$p = 1.0$ & 89.27 & 89.12 & 87.58 & 89.03 \\
\Xhline{4\arrayrulewidth}
\end{tabular}
\caption{The accuracy (in \%) of vanilla trained models on the CIFAR10 test set under various settings, including \textit{Signed Kaiming Constant} (Signed KC) and the binary initialization. We include models both with and without the last batch normalization layer (LBN). The best results are marked in bold.}
\label{tbl:vanilla_ablation}
\end{table}

\subsubsection{More results of Baselines}
\label{subsec: more baseline results}

We show in Table~\ref{tbl:complete comparison} the complete set of experiments of baseline algorithms \cliucamera{on CIFAR10 and CIFAR100} as a complementary of Table~\ref{tbl:main comparison}.
\change{Specifically, we compare baselines with different architectures and with different pruning strategies.
First, the last batch normalization layer (LBN) does not improve the baselines that update model parameters in the full-precision setting, because the magnitude of the output logits can be automatically adjusted in these cases.
There is no need to insert another normalization layer.
\zzhao{For FlyingBird(+), BCS and HYDRA, adding LBN to a binary network will most likely be beneficial to a better performance. This observation is consistent with our claim in Appendix~\ref{subsec:normalization}. As for ATMC, it is actually not pruning a truly binary network since the value of model parameters are trainable and not necessarily $+1$ or $-1$, so adding LBN might not be useful in this case. }
For the pruning strategy, \textit{adaptive pruning} strategy with $p = 0.1$ always has better performance \zzhao{than} the \textit{fixed pruning rate} strategy, i.e., $p = 1.0$.
This is because the pruning rate here is very high $r = 0.99$, and we need a small value of $p$ based on the analysis in Section~\ref{subsec:adaptivepruning}.
Furthermore, we provide the performance of TRADES~\cite{zhang2019theoretically}, which trades clean accuracy for adversarial accuracy.
Compared with adversarial training (AT), TRADES achieves competitive performance in the full precision cases, but it performance degrades significantly in the binary cases.
}

\begin{table*}[h!]
\small
\centering
\begin{tabular}{p{1.9cm}p{2.3cm}<{\centering}p{1.6cm}<{\centering}p{0.8cm}<{\centering}p{0.8cm}<{\centering}p{0.8cm}<{\centering}p{0.8cm}<{\centering}}
\Xhline{4\arrayrulewidth}
\multirow{2}{*}{Method} & \multirow{2}{*}{Architecture} & Pruning & \multicolumn{2}{c}{CIFAR10} & \multicolumn{2}{c}{CIFAR100} \\
& & Strategy & FP & Binary & FP & Binary \\
\hline
AT    & RN34 & Not Pruned          & 43.26 & 40.34 & 36.63 & 26.49 \\
AT    & RN34-LBN & Not Pruned          & 42.39 & 39.58 & 35.15 & 32.98 \\
TRADES    & RN34 & Not Pruned          & 49.07 & 30.18 & 35.28 & 29.64 \\
TRADES    & RN34-LBN & Not Pruned    & 48.27 & 37.91 & 31.23 & 31.26 \\
\hdashline
FlyingBird & RN34 & Dynamic & \underline{45.86} & 34.37 & \underline{35.91} & 22.49 \\
FlyingBird+ & RN34 & Dynamic & 44.57 & 33.33 & 34.30 & 22.64 \\
FlyingBird & RN34-LBN & Dynamic & 45.58 & 37.18 & 35.06 & 24.94 \\
FlyingBird+ & RN34-LBN & Dynamic & 44.44 & 37.48 & 34.03 & 24.50 \\
BCS & RN34 & Dynamic & 43.51 & 22.61 & 31.85 & 11.96 \\
BCS & RN34-LBN & Dynamic & 42.02 & 30.67 & 31.16 & \zzhaocamera{17.97}\\ 
RST & RN34 & $p=1.0$ & 34.95 & - & 21.96 & - \\
RST & RN34-LBN & $p=1.0$ & 37.23 & - & 23.14 & - \\
HYDRA & RN34 & $p = 0.1$         & 42.73 & 29.28 & 33.00 & 23.60 \\
HYDRA & RN34 & $p = 1.0$         & 40.51 & 26.40 & 31.09 & 18.24 \\
HYDRA & RN34-LBN & $p = 0.1$      & 40.55 & 33.99 & 13.63 & 25.53 \\
HYDRA & RN34-LBN & $p = 1.0$      & 32.93 & 26.23 & 29.96 & 18.91 \\
ATMC  & RN34 & Global            & 34.14 & 25.62 & 25.10 & 11.09 \\
ATMC  & RN34 & $p = 0.1$         & 34.58 & 24.65 & 25.37 & 11.04 \\
ATMC  & RN34 & $p = 1.0$         & 30.50 & 20.21 & 22.28 & 2.53  \\
ATMC  & RN34-LBN & Global         & 33.55 & 19.01 & 23.16 & 15.73 \\
ATMC  & RN34-LBN & $p = 0.1$      & 31.61 & 22.88 & 25.16 & 17.33 \\
ATMC  & RN34-LBN & $p = 1.0$      & 27.88 & 13.22 & 22.12 & 9.55  \\
AT & Small RN34-p0.1 & Not Pruned  & 42.01 & 32.54 & 28.46 & 16.18 \\
AT & Small RN34-p1.0 & Not Pruned  & 38.81 & 26.03 & 27.68 & 15.85 \\
TRADES & Small RN34-p0.1 & Not Pruned  & 42.60 & 29.92 & 28.44 & 15.25 \\
TRADES & Small RN34-p1.0 & Not Pruned  & 38.53 & 24.83 & 27.63 & 13.16 \\
Ours & RN34-LBN & $p = 0.1$      & -     & \textbf{45.06} & -     & \textbf{34.83} \\
Ours & RN34-LBN & $p = 1.0$      & -     & 34.57 & -     & 26.32 \\
Ours (fast) & RN34-LBN & $p = 0.1$ & -     & 40.77 & -     & 34.45 \\
Ours (fast) & RN34-LBN & $p = 1.0$ & -     & 29.68 & -     & 24.97 \\
\Xhline{4\arrayrulewidth}
\end{tabular}
\caption{Robust accuracy (in \%) on the CIFAR10 and CIFAR100 test sets for AT, TRADES, FlyingBird(+), BCS, RST, HYDRA, ATMC and our proposed method. ``RN34-LBN'' represents RN34 with the last batch normalization layer. 
\zzhao{``Small RN34'' here refers to Small RN34-p0.1 in Table~\ref{tbl:smallnet} of Appendix~\ref{subsec:exp_setting}.}
Among the compressed models, the best results for full precision (FP) models are underlined; the best results for binary models are marked in bold.} 
\label{tbl:complete comparison}
\end{table*}

\subsubsection{Clean accuracy of Models in Table~\ref{tbl:main comparison}} \label{subsec:app_vanilla_accuracy}

Table~\ref{tbl:comparison vanilla} shows the \cliucamera{accuracy on the clean test set} of the models in Table~\ref{tbl:main comparison}. 
In the CIFAR10 dataset, our pruned networks with both normal and fast pruning achieve the highest vanilla accuracy among all binary networks.
Although the accuracy is lower than full-precision networks by ATMC, our model performs notably better ($>10\%$) under AutoAttack.
In the CIFAR100 dataset, our model using FGSM with ATTA has the best vanilla accuracy among both full-precision networks and binary networks, and also achieves comparable robust accuracy to them, as shown in Table~\ref{tbl:main comparison}.
Our model using PGD also achieves competitive performance, better than all other binary networks.
\zzhaocamera{In the ImageNet100 dataset, our model still outperforms all other pruned binary models, although it is worse than some full precision models.}
These results indicate that our models can achieve competitive robust accuracy without losing too much vanilla accuracy, hence more powerful in real applications where both robust and vanilla accuracy are important.

\begin{table*}[!ht]
\centering
\begin{tabular}{p{1.7cm}p{2.1cm}<{\centering}p{1.6cm}<{\centering}p{0.8cm}<{\centering}p{0.8cm}<{\centering}p{0.8cm}<{\centering}p{0.8cm}<{\centering}p{0.8cm}<{\centering}p{0.8cm}<{\centering}}
\Xhline{4\arrayrulewidth}
\multirow{2}{*}{Method} & \multirow{2}{*}{Architecture} & Pruning & \multicolumn{2}{c}{CIFAR10} & \multicolumn{2}{c}{CIFAR100} & \multicolumn{2}{c}{ImageNet100}\\
& & Strategy & FP & Binary & FP & Binary & FP & Binary\\
\hline
AT & RN34       & Not Pruned & 80.99 & 74.37 & 61.48 & 47.87 & 78.98 & 63.76 \\
AT & RN34-LBN   & Not Pruned & 80.96 & 74.17 & 57.73 & 60.08 & 77.66 & 64.60 \\
\hdashline
AT & Small RN34 & Not Pruned & 74.76 & 58.69 & 52.77 & 28.81 & 49.64 & 21.12 \\
FlyingBird  & RN34 & Dynamic & 79.29 & 62.28 & \underline{62.12} & 43.66 & 66.66 & 19.74 \\
FlyingBird+ & RN34 & Dynamic & 77.01 & 62.69 & 59.09 & 41.69 & 66.66 & 19.74 \\
BCS         & RN34 & Dynamic & 74.75 & - & 53.82 & - & - & - \\
RST & RN34     & $p = 1.0$ & 65.93 & - & 38.87 & - & 42.70 & - \\
RST & RN34-LBN & $p = 1.0$ & 67.45 & - & 42.95 & - & 46.22 & - \\
HYDRA & RN34   & $p = 0.1$ & 75.31 & 62.09 & 55.92 & 45.96 & \underline{67.76} & 33.18 \\
ATMC  & RN34   & Global    & \underline{81.85} & 72.97 & 57.15 & 36.39 & 60.68 & 26.80 \\
ATMC  & RN34   & $p = 0.1$ & 81.37 & 73.34 & 59.99 & 32.68 & 61.88 & 16.34 \\
Ours       & RN34-LBN & $p = 0.1$ & - & 76.59 & - & 60.16 & \multirow{2}{*}{-} & \multirow{2}{*}{\textbf{58.94}} \\
Ours(fast) & RN34-LBN & $p = 0.1$ & - & \textbf{81.63} & - & \textbf{63.73} \\
\Xhline{4\arrayrulewidth}
\end{tabular}
\caption{The accuracy (in \%) on the clean inputs of the methods studied in Section~\ref{subsec:compare_existing}. ``RN34-LBN'' represents RN34 with the last batch normalization layer. Among the pruned models, the best results in the full precision (FP) cases are underlined and the best results in the binary cases are marked in bold.} \label{tbl:comparison vanilla}
\end{table*}

\subsubsection{\nipsedit{Our Method in the Non-adversarial Cases}} \label{subsec:nonadversarial}

\nipsedit{Vanilla training can be considered as a special case of adversarial training: the case when $\epsilon = 0$.
Therefore, our methods, as well as baselines, are applicable to vanilla training. The results of the cases when $\epsilon = 0$ are demonstrated in Table~\ref{tbl:vanilla}.
Since there are no adversarial attacks in vanilla training, the acceleration used in ``Ours (fast)'' is not applicable here.
The results in Table~\ref{tbl:vanilla} demonstrate the consistent observations with Table~\ref{tbl:main comparison}: our proposed methods achieve the best performance among binary networks.}

\begin{table*}[!ht]
\centering
\begin{tabular}{p{1.7cm}p{2.1cm}<{\centering}p{1.6cm}<{\centering}p{0.8cm}<{\centering}p{0.8cm}<{\centering}p{0.8cm}<{\centering}p{0.8cm}<{\centering}p{0.8cm}<{\centering}p{0.8cm}<{\centering}}
\Xhline{4\arrayrulewidth}
\multirow{2}{*}{Method} & \multirow{2}{*}{Architecture} & Pruning & \multicolumn{2}{c}{CIFAR10} & \multicolumn{2}{c}{CIFAR100} & \multicolumn{2}{c}{ImageNet100}\\
& & Strategy & FP & Binary & FP & Binary & FP & Binary\\
\hline
AT          & RN34      & Not Pruned    & 94.80    & 90.11     & 76.39 & 70.02 & 80.26 & 68.26\\
AT          & RN34-LBN  & Not Pruned    & 94.79    & 92.46     & 76.85 & 73.49 & 79.84 & 73.88\\
\hdashline
AT          & Small RN34& Not Pruned    & 91.99    & 85.61     & 65.48 & 43.46 & 58.14 & 29.62\\
FlyingBird  & RN34      & Dynamic& \underline{93.41}& 88.96    & 71.77 & 61.50 & 74.06 & 26.06  \\
FlyingBird+ & RN34      & Dynamic       &92.28&86.44&\underline{72.03} & 58.09 & 74.40 & 27.52 \\
BCS         & RN34      & Dynamic       & 90.69    & -         & 67.39 & - & - & -\\
RST         & RN34      & $p = 1.0$     & 88.43    & -         & 56.65 & - & 50.18 & -\\
RST         & RN34-LBN  & $p = 1.0$     & 89.14    & -         & 62.93 & - & 61.52 & -\\
HYDRA       & RN34      & $p = 0.1$     & 91.13    & 88.10     & 68.84 & 62.10 & \underline{76.42} & 49.40 \\
ATMC        & RN34      & Global        & 92.01    & 88.40     & 67.45 & 51.96 & 69.36 & 35.30 \\
ATMC        & RN34      & $p = 0.1$     & 91.32    & 79.46     & 68.03 & 50.94 & 70.12 & 33.52 \\
Ours        & RN34-LBN  & $p = 0.1$     & -    &\textbf{93.99} & -     & \textbf{75.37} & - & \textbf{72.80}\\
\Xhline{4\arrayrulewidth}
\end{tabular}
\caption{\nipsedit{Clean accuracy (in \%) on the CIFAR10, CIFAR100 and ImageNet100 test sets for the baselines and our proposed method in the non-adversarial case, i.e., $\epsilon = 0$. ``RN34-LBN'' represents ResNet34 with the last batch normalization layer. ``Small RN34'' refers to Small RN34-p0.1 in Table~\ref{tbl:smallnet} of Appendix~\ref{subsec:exp_setting}. The pruning rate is set to $0.99$ except for the not-pruned methods. \cliucamera{Among the pruned models}, the best results for the full-precision (FP) models are underlined; the best results for the binary models are marked in bold.}}~\label{tbl:vanilla}
\end{table*}

\subsubsection{Mask of the Pruned Network} \label{subsec:mask}

We have demonstrated that the masks of the pruned network obtained by our method are structured to some degree in Section~\ref{subsec:main subnetwork pattern}. We have also analyzed the structure of a randomly pruned network in Appendix~\ref{subsec:random structure}.

Figure~\ref{fig:mask_full} shows the one of the convolutional layers in our pruned RN34 network. 
\zzhaocamera{We resize the layer parameters as grids of shape $(r_{out}, r_{in})$ for visualization. Each grid represents a 3-by-3 kernel. So the shape of parameters is $(r_{out} \times 3, r_{in} \times 3)$. The retained parameters in each kernel are marked in blue.}
The pruning rate for this layer is $r = 0.99$.
We highlight the input channels that are totally pruned in orange.
We also use a white bar at the top of the figure to indicate these empty input channels.

\begin{figure}[htb]
    \centering
    \includegraphics[width=0.9\textwidth]{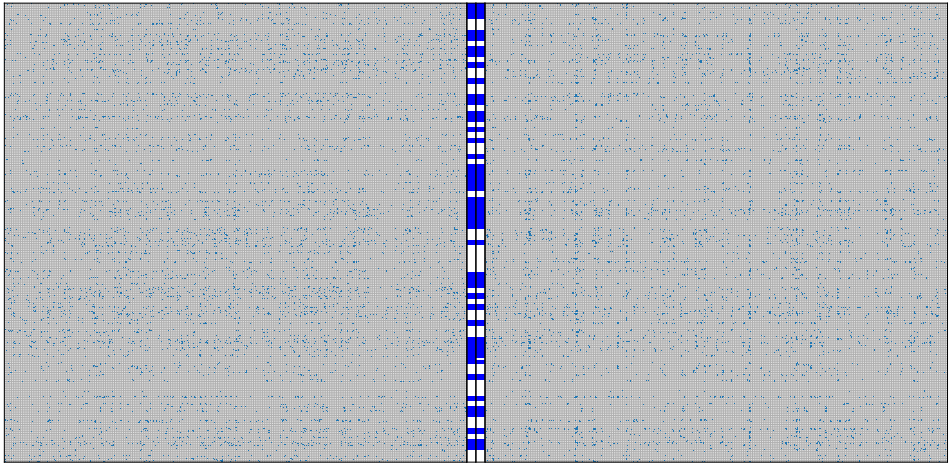}
    \caption{Distribution of weights in two consecutive layers. In layer1 (left), the masks are reshaped into ($r_{out}$ $\times$ 3, $r_{in}$ $\times$ 3) while masks in layer2 (right) are reshaped into ($r_{in}$ $\times$ 3, $r_{out}$ $\times$ 3). The output channels totally pruned in layer1 and the input channels totally pruned in layer2 are highlighted as the white bars in the middle. \cliucamera{Due to the large number of parameters in these layers, readers could zoom in this figure to see more details.}}
    \label{fig:mask 2 layers}
\end{figure}

In Section~\ref{subsec:main subnetwork pattern}, we also point out the aligned pruning pattern in the two consecutive layers, layer1 and layer2, of the same residual block in RN34.
Figure~\ref{fig:mask 2 layers} shows their pruning masks. 
The side bars show which channel is non-empty(colored in blue). 
For convenience, layer1 is resized in $(r_{out} \times 3, r_{in} \times 3)$, and layer2 is organized in $(r_{in} \times 3, r_{out} \times 3)$. It is interesting that the pruned input channels of layer2 are well aligned with the pruned output channels of layer1.

Note that our finding also holds in the vanilla settings, i.e. pruning with clean examples. We think this observation enables a possible way for regular pruning.

\subsubsection{Learning Curves of Adaptive Pruning with Different $p$ Values} \label{subsec:p_sensitivity_analysis}

\begin{figure}[!ht]
    \centering
    \begin{subfigure}{0.22\textwidth}
        \centering
        \includegraphics[width=\linewidth]{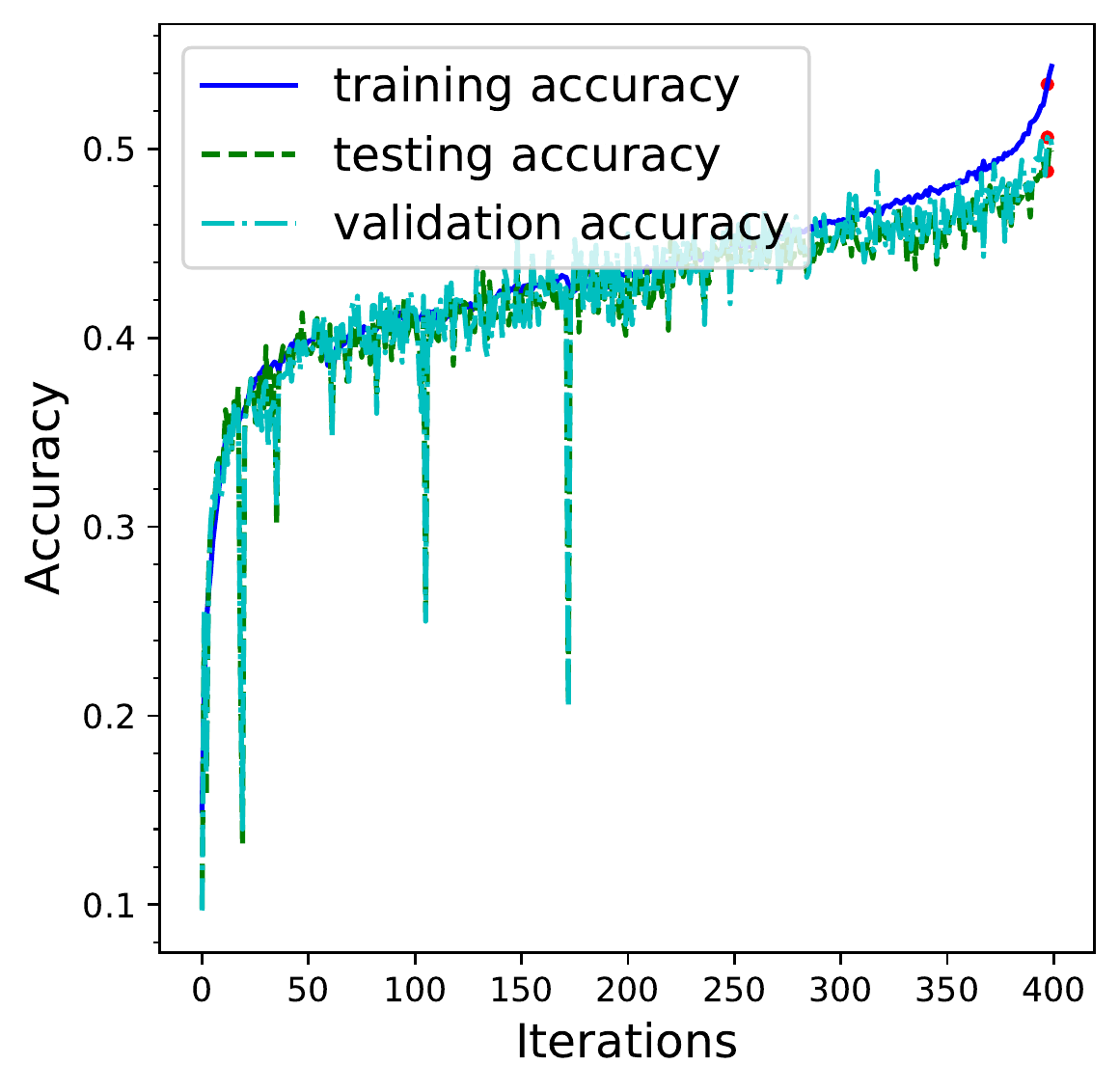}
        \caption{$p=0.1$}
    \end{subfigure}
    \begin{subfigure}{0.22\textwidth}
        \centering
        \includegraphics[width=\linewidth]{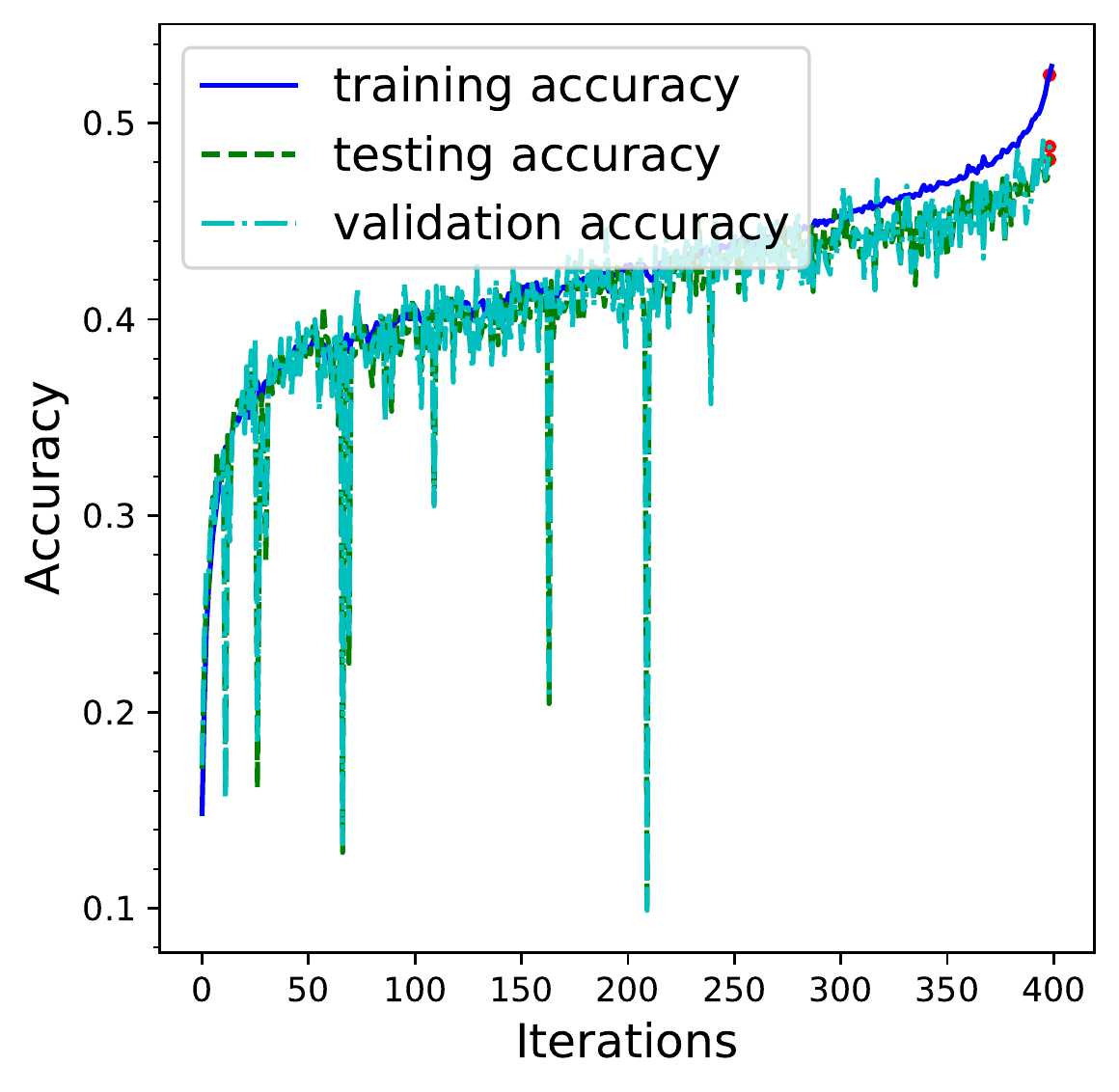}
        \caption{$p=0.5$}
    \end{subfigure}
    \begin{subfigure}{0.22\textwidth}
        \centering
        \includegraphics[width=\linewidth]{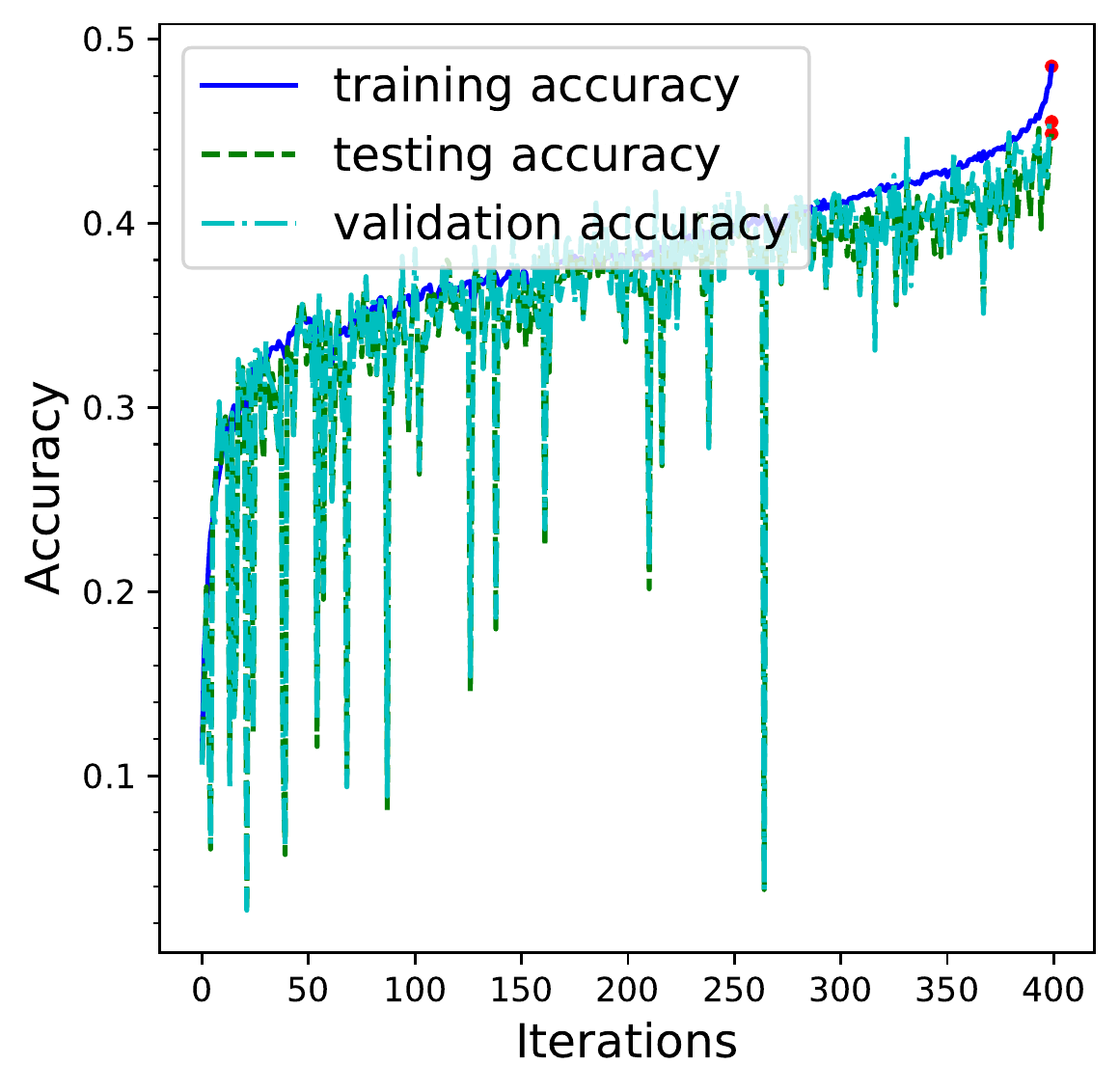}
        \caption{$p=0.8$}
    \end{subfigure}
    \begin{subfigure}{0.22\textwidth}
        \centering
        \includegraphics[width=\linewidth]{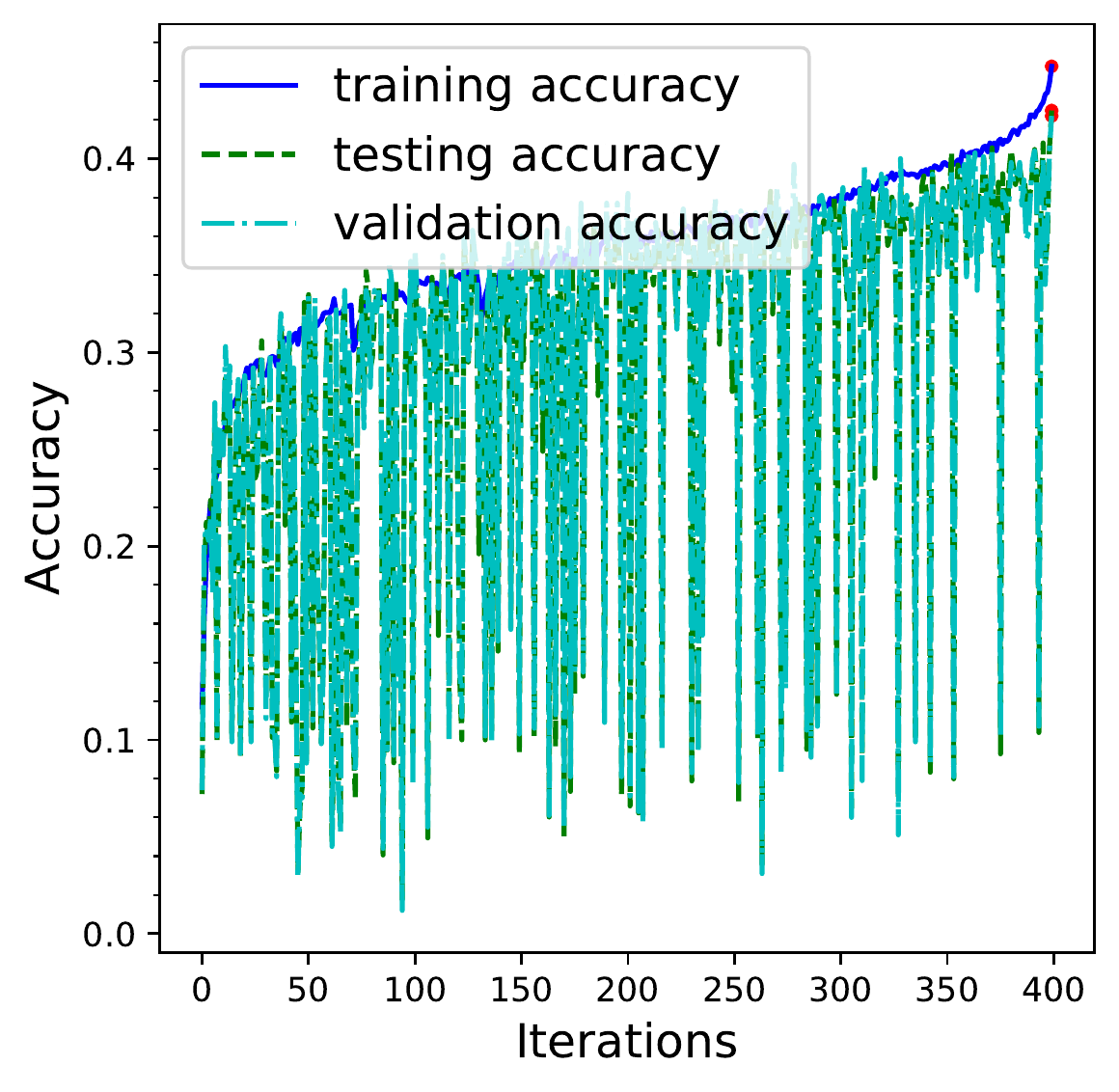}
        \caption{$p=1.0$}
    \end{subfigure}
    
    \begin{subfigure}{0.22\textwidth}
        \centering
        \includegraphics[width=\linewidth]{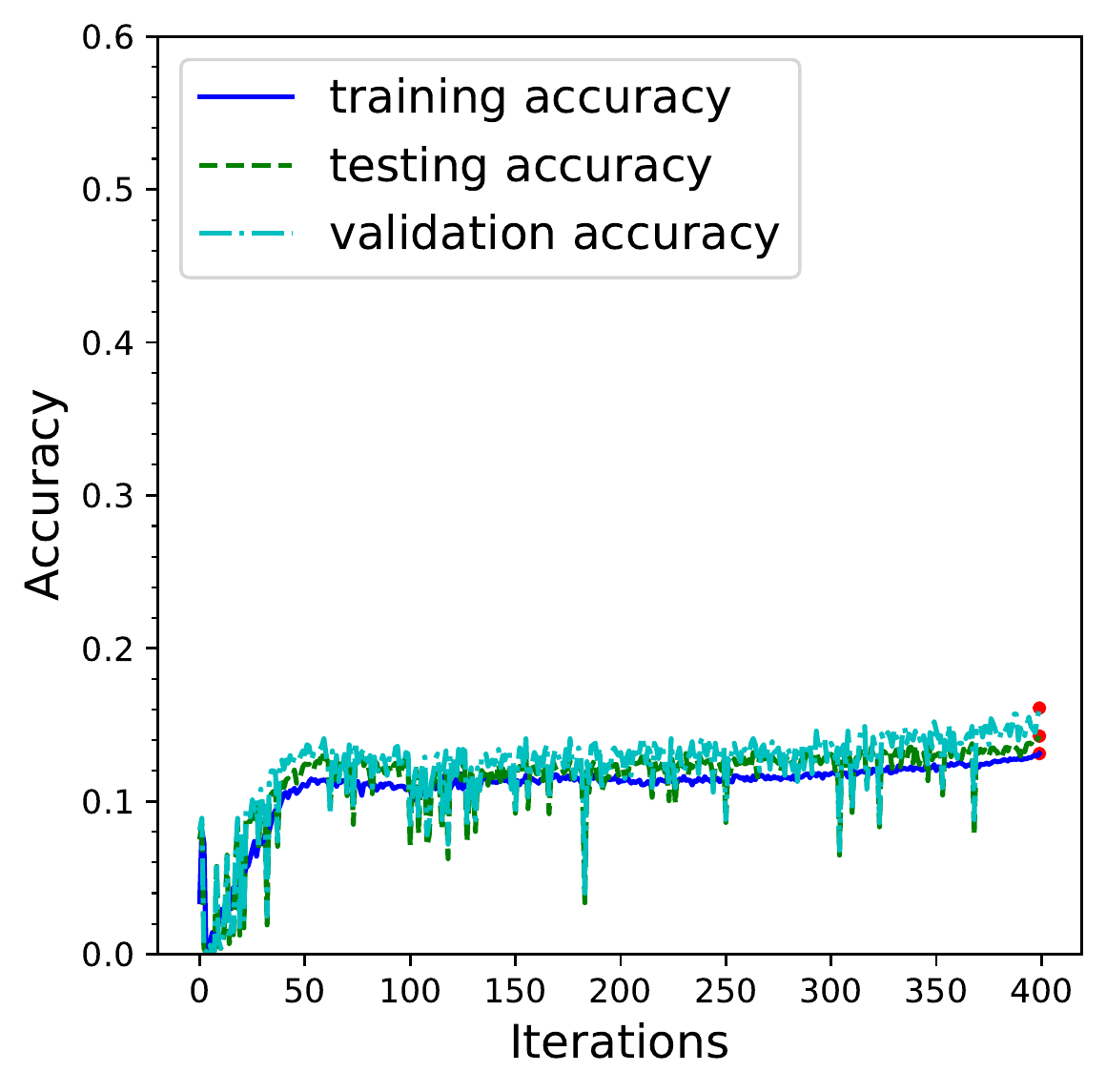}
        \caption{$p=0.1$}
    \end{subfigure}
    \begin{subfigure}{0.22\textwidth}
        \centering
        \includegraphics[width=\linewidth]{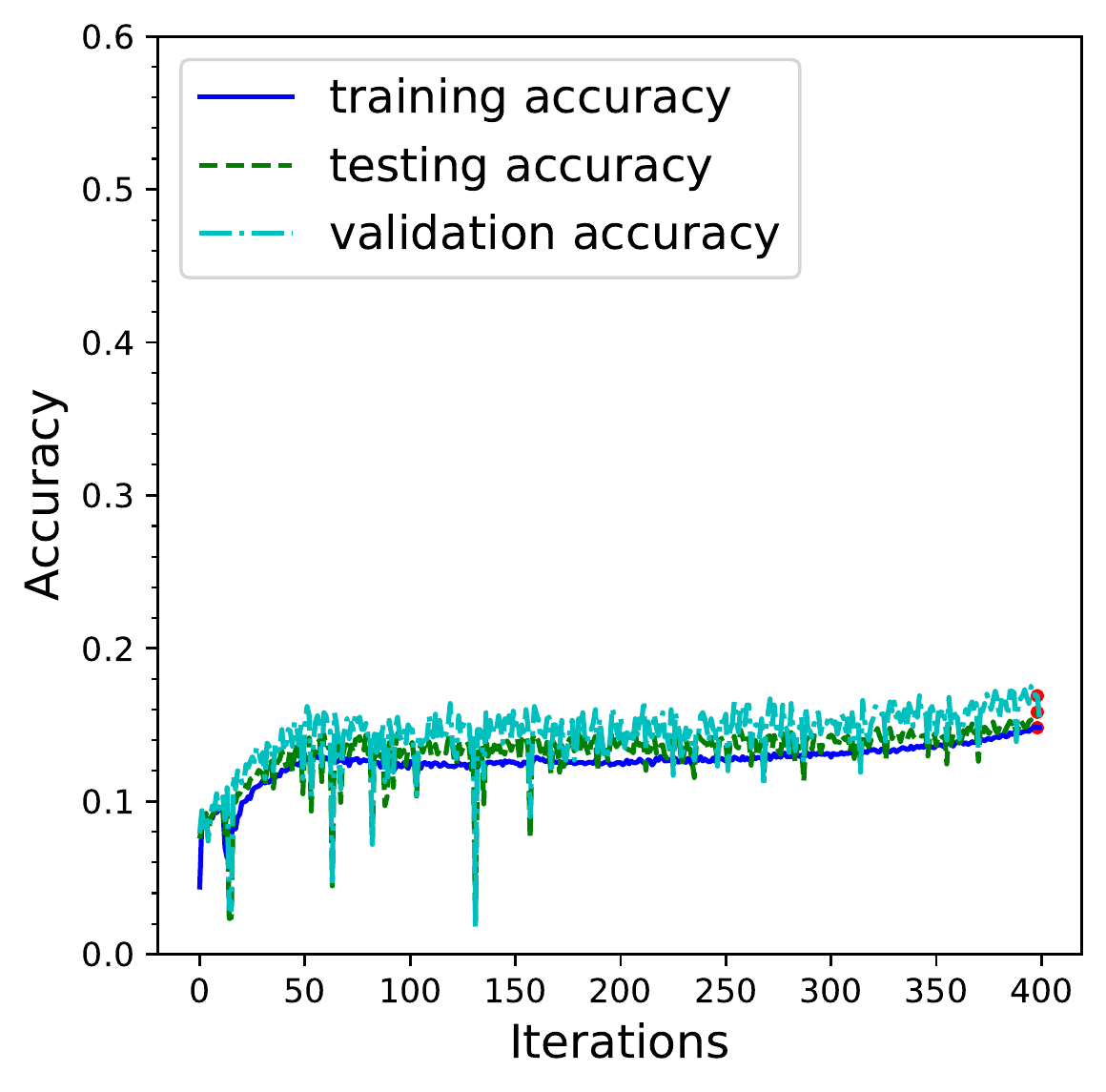}
        \caption{$p=0.5$}
    \end{subfigure}
    \begin{subfigure}{0.22\textwidth}
        \centering
        \includegraphics[width=\linewidth]{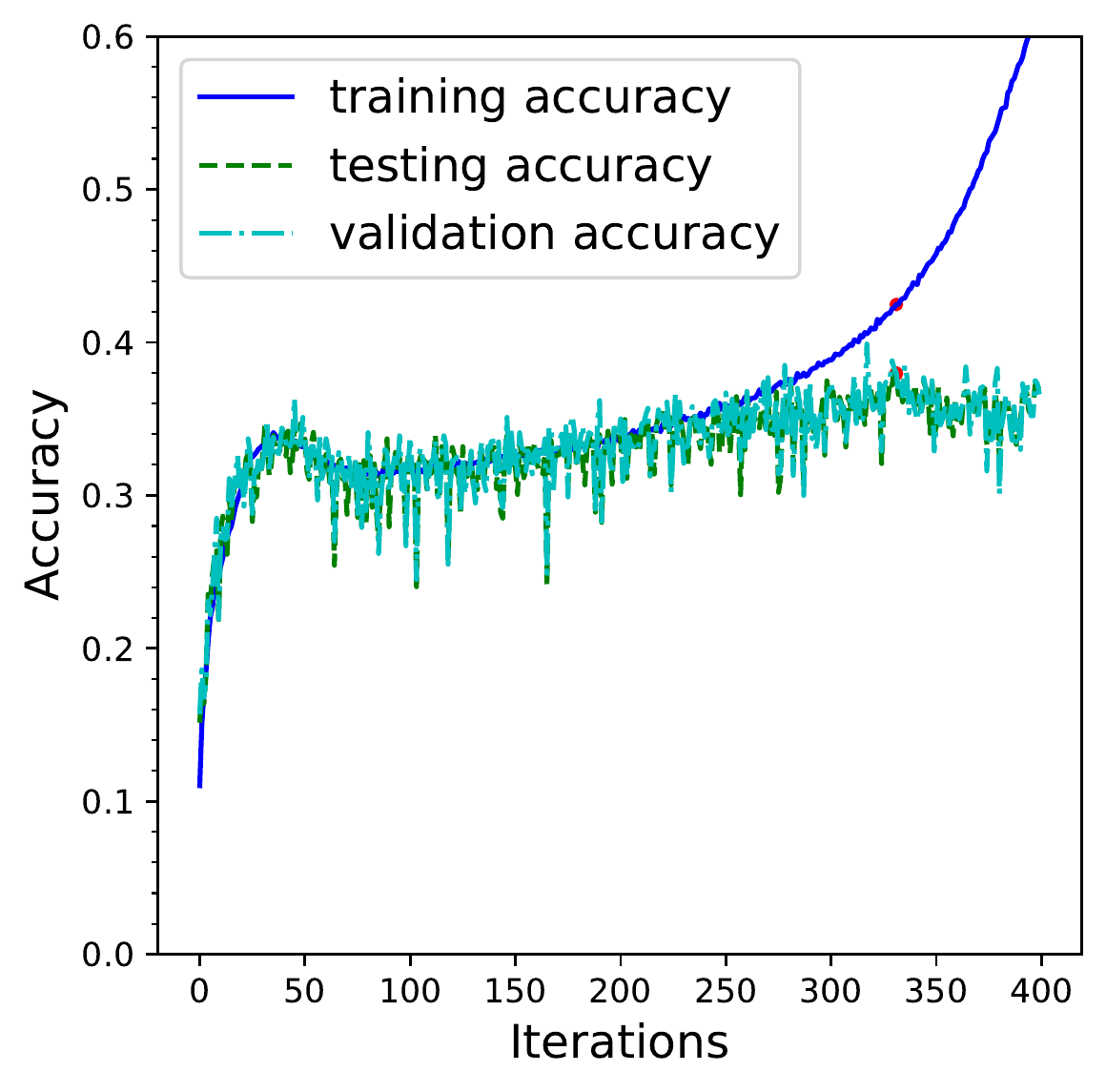}
        \caption{$p=0.8$}
    \end{subfigure}
    \begin{subfigure}{0.22\textwidth}
        \centering
        \includegraphics[width=\linewidth]{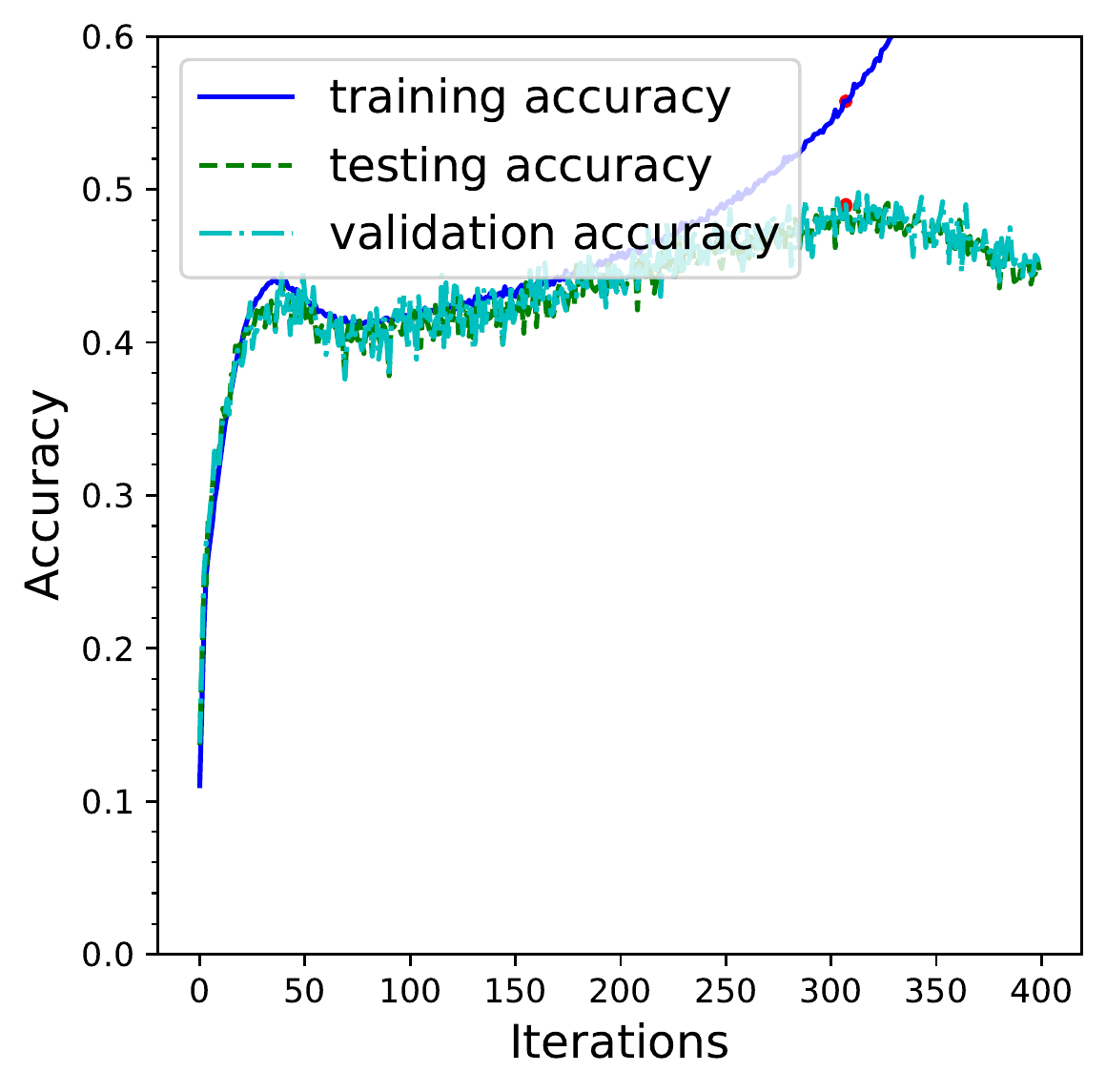}
        \caption{$p=1.0$}
    
    \end{subfigure}
    \caption{Learning curves of our proposed method under adaptive pruning strategy with different values of $p$. The pruning ratio is $0.99$ for figure (a) - (d) and is $0.5$ for figure (e) - (h).}
    \label{fig:stability}
\end{figure}

We plot the learning curves when we use the \textit{adaptive pruning} strategy with different values of $p$ in Figure~\ref{fig:stability}.
Here, we use $r = 0.99$ and $r = 0.5$ as two examples.
Based on the results of Table~\ref{tbl:sp_ratio}, our method achieves the best performance under $p = 0.1$ when $r= 0.99$ and under $p = 1.0$ when $r = 0.5$.
The learning curves in Figure~\ref{fig:stability} indicate that the training process is quite unstable when using the inappropriate pruning strategy, leading to suboptimal performance.



\begin{figure*}[htb]
    \centering
    \includegraphics[width=\textwidth]{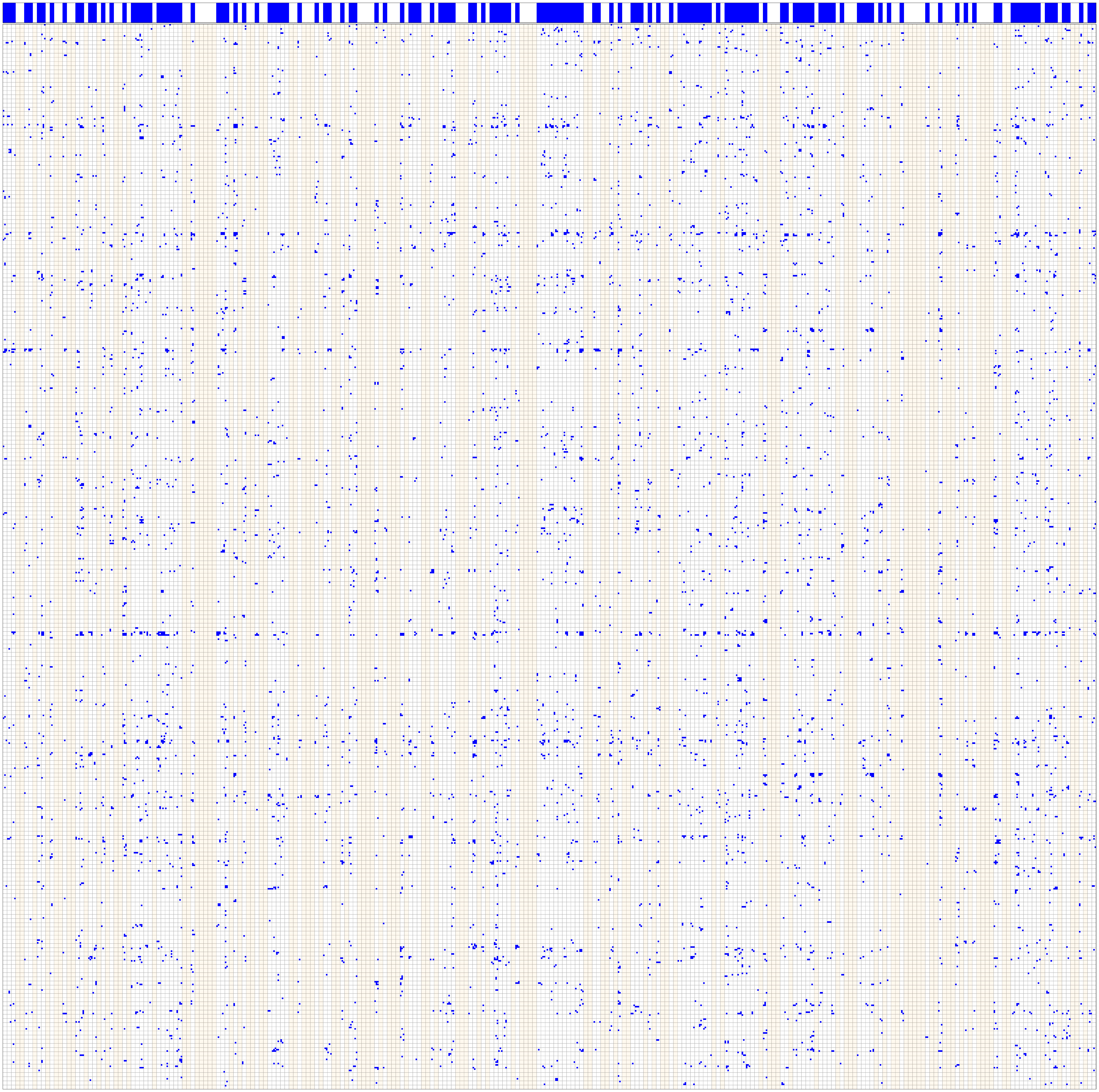}
    \caption{Mask visualization of the weight of a random convolutional layer in our model. The parameters retained is highlighted as blue dots. The dimension of the convolutional kernel is ($r_{out}$, $r_{in}$, 3, 3). We reshape this kernel in rectangle of shape ($r_{out}$ $\times$ 3, $r_{in}$ $\times$ 3). Channels with no remaining weight are colored orange. The top bar indicates whether the channel is empty (white) or not (blue). Due to the large number of parameters in this layer, readers could zoom in this figure to see more details.}
    \label{fig:mask_full}
\end{figure*}

\end{document}